%% file: main.tex
\documentclass[10pt,journal,compsoc]{IEEEtran}
\ifCLASSOPTIONcompsoc
  \usepackage[nocompress]{cite}
\else
  \usepackage{cite}
\fi
\ifCLASSINFOpdf
  \usepackage[pdftex]{graphicx}
\else
\fi

\usepackage{url}
\usepackage{times}
\usepackage{epsfig}
\usepackage{graphicx}
\usepackage{amsmath}
\usepackage{amssymb}
\usepackage{wrapfig}
\usepackage{multirow}
\usepackage{color}

\usepackage{caption}
\DeclareCaptionType{noticebox}
\usepackage{subcaption}

\usepackage{hyperref}

\usepackage{tikz}
\usepackage{textcomp}
\usepackage{lipsum}

\newcommand\copyrighttext{%
  \footnotesize \textcopyright 2019 IEEE. Personal use of this material is permitted.
  Permission from IEEE must be obtained for all other uses, in any current or future
  media, including reprinting/republishing this material for advertising or promotional
  purposes, creating new collective works, for resale or redistribution to servers or
  lists, or reuse of any copyrighted component of this work in other works.
  DOI: {10.1109/TPAMI.2019.2918284}}
  
\newcommand\copyrightnotice{%
\begin{tikzpicture}[remember picture,overlay]
\node[anchor=south,yshift=10pt] at (current page.south) {\fbox{\parbox{\dimexpr\textwidth-\fboxsep-\fboxrule\relax}{\copyrighttext}}};
\end{tikzpicture}%
}

\begin{document}

\input{macros}
% paper title
% Titles are generally capitalized except for words such as a, an, and, as,
% at, but, by, for, in, nor, of, on, or, the, to and up, which are usually
% not capitalized unless they are the first or last word of the title.
% Linebreaks \\ can be used within to get better formatting as desired.
% Do not put math or special symbols in the title.
% \title{Densely Connected Convolutional Networks}
\title{Convolutional Networks with Dense Connectivity}
%
%
% author names and IEEE memberships
% note positions of commas and nonbreaking spaces ( ~ ) LaTeX will not break
% a structure at a ~ so this keeps an author's name from being broken across
% two lines.
% use \thanks{} to gain access to the first footnote area
% a separate \thanks must be used for each paragraph as LaTeX2e's \thanks
% was not built to handle multiple paragraphs
%
%
%\IEEEcompsocitemizethanks is a special \thanks that produces the bulleted
% lists the Computer Society journals use for "first footnote" author
% affiliations. Use \IEEEcompsocthanksitem which works much like \item
% for each affiliation group. When not in compsoc mode,
% \IEEEcompsocitemizethanks becomes like \thanks and
% \IEEEcompsocthanksitem becomes a line break with idention. This
% facilitates dual compilation, although admittedly the differences in the
% desired content of \author between the different types of papers makes a
% one-size-fits-all approach a daunting prospect. For instance, compsoc
% journal papers have the author affiliations above the "Manuscript
% received ..."  text while in non-compsoc journals this is reversed. Sigh.

\author{Gao~Huang, Zhuang~Liu, Geoff Pleiss, Laurens van der Maaten and Kilian Q. Weinberger
        %and~Jane~Doe,~\IEEEmembership{Life~Fellow,~IEEE}% <-this % stops a space

\IEEEcompsocitemizethanks{
\IEEEcompsocthanksitem Gao Huang is with the Department
of Automation, Tsinghua University, Beijing, 100084. (E-mail: gaohuang@tsinghua.edu.cn).
%\protect\\
% note need leading \protect in front of \\ to get a newline within \thanks as
% \\ is fragile and will error, could use \hfil\break instead.

\IEEEcompsocthanksitem Geoff Pleiss and Kilian Q. Weinberger are with the Department
of Computer Science, Cornell University, Ithaca,
NY, 14850. (E-mail: kwq4@cornell.edu, geoff@cs.cornell.edu).
\IEEEcompsocthanksitem Zhuang Liu is with Berkeley Artificial Intelligence Research, UC Berkeley, Berkeley, CA 94704. (E-mail: zhuangl@berkeley.edu).
\IEEEcompsocthanksitem Laurens van der Maaten is with Facebook AI Research. (E-mail: lvdmaaten@fb.com).

}

% <-this % stops an unwanted space

%\thanks{Manuscript received April 19, 2005; revised August 26, 2015.}
}

% note the % following the last \IEEEmembership and also \thanks -
% these prevent an unwanted space from occurring between the last author name
% and the end of the author line. i.e., if you had this:
%
% \author{....lastname \thanks{...} \thanks{...} }
%                     ^------------^------------^----Do not want these spaces!
%
% a space would be appended to the last name and could cause every name on that
% line to be shifted left slightly. This is one of those "LaTeX things". For
% instance, "\textbf{A} \textbf{B}" will typeset as "A B" not "AB". To get
% "AB" then you have to do: "\textbf{A}\textbf{B}"
% \thanks is no different in this regard, so shield the last } of each \thanks
% that ends a line with a % and do not let a space in before the next \thanks.
% Spaces after \IEEEmembership other than the last one are OK (and needed) as
% you are supposed to have spaces between the names. For what it is worth,
% this is a minor point as most people would not even notice if the said evil
% space somehow managed to creep in.

% The paper headers
\markboth{IEEE Transactions on Pattern Analysis and Machine Intelligence}%
{Shell \MakeLowercase{\textit{et al.}}: Bare Demo of IEEEtran.cls for Computer Society Journals}
% The only time the second header will appear is for the odd numbered pages
% after the title page when using the twoside option.
%
% *** Note that you probably will NOT want to include the author's ***
% *** name in the headers of peer review papers.                   ***
% You can use \ifCLASSOPTIONpeerreview for conditional compilation here if
% you desire.

% The publisher's ID mark at the bottom of the page is less important with
% Computer Society journal papers as those publications place the marks
% outside of the main text columns and, therefore, unlike regular IEEE
% journals, the available text space is not reduced by their presence.
% If you want to put a publisher's ID mark on the page you can do it like
% this:
%\IEEEpubid{0000--0000/00\$00.00~\copyright~2015 IEEE}
% or like this to get the Computer Society new two part style.
%\IEEEpubid{\makebox[\columnwidth]{\hfill 0000--0000/00/\$00.00~\copyright~2015 IEEE}%
%\hspace{\columnsep}\makebox[\columnwidth]{Published by the IEEE Computer Society\hfill}}
% Remember, if you use this you must call \IEEEpubidadjcol in the second
% column for its text to clear the IEEEpubid mark (Computer Society jorunal
% papers don't need this extra clearance.)

% use for special paper notices
%\IEEEspecialpapernotice{(Invited Paper)}

%\renewcommand\paragraph{\@startsection{paragraph}{4}{\z@}%
%  {-3.25ex \@plus -1ex \@minus -0.2ex}%
%  {0.01pt}%
%  {\raggedsection\normalfont\sectfont\nobreak\size@paragraph}%
%}

\newcommand{\para}[1]{\textbf{#1}}
%\renewcommand{\paragraph}[1]{\textbf{#1}}

% for Computer Society papers, we must declare the abstract and index terms
% PRIOR to the title within the \IEEEtitleabstractindextext IEEEtran
% command as these need to go into the title area created by \maketitle.
% As a general rule, do not put math, special symbols or citations
% in the abstract or keywords.
\IEEEtitleabstractindextext{%
\begin{abstract}
Recent work has shown that convolutional networks can be substantially deeper, more accurate, and efficient to train if they contain shorter connections between layers close to the input and those close to the output.
In this paper, we embrace this observation and introduce the \methodnamecap{} (\methodnameshort{}), which connects each layer to every other layer in a feed-forward fashion. 
Whereas traditional convolutional networks with $L$ layers have $L$ connections---one between each layer and its subsequent layer---our network has $\frac{L(L+1)}{2}$ direct connections.
For each layer, the feature-maps of all preceding layers are used as inputs, and its own feature-maps are used as inputs into all subsequent layers.
\methodnameshorts{} have several compelling advantages:
they alleviate the vanishing-gradient problem, encourage feature reuse and substantially improve parameter efficiency.
We evaluate our proposed architecture on four highly competitive object recognition benchmark tasks (CIFAR-10, CIFAR-100, SVHN, and ImageNet). \methodnameshorts{} obtain significant improvements over the state-of-the-art on most of them, whilst requiring less parameters and computation to achieve high performance.
%Code and models are available at \url{https://github.com/liuzhuang13/DenseNet}.
\end{abstract}

% Note that keywords are not normally used for peerreview papers.
}

% make the title area
\maketitle

\copyrightnotice

% To allow for easy dual compilation without having to reenter the
% abstract/keywords data, the \IEEEtitleabstractindextext text will
% not be used in maketitle, but will appear (i.e., to be "transported")
% here as \IEEEdisplaynontitleabstractindextext when the compsoc
% or transmag modes are not selected <OR> if conference mode is selected
% - because all conference papers position the abstract like regular
% papers do.
\IEEEdisplaynontitleabstractindextext
% \IEEEdisplaynontitleabstractindextext has no effect when using
% compsoc or transmag under a non-conference mode.

% For peer review papers, you can put extra information on the cover
% page as needed:
% \ifCLASSOPTIONpeerreview
% \begin{center} \bfseries EDICS Category: 3-BBND \end{center}
% \fi
%
% For peerreview papers, this IEEEtran command inserts a page break and
% creates the second title. It will be ignored for other modes.
\IEEEpeerreviewmaketitle

\IEEEraisesectionheading{\section{Introduction}\label{sec:introduction}}
% Computer Society journal (but not conference!) papers do something unusual
% with the very first section heading (almost always called "Introduction").
% They place it ABOVE the main text! IEEEtran.cls does not automatically do
% this for you, but you can achieve this effect with the provided
% \IEEEraisesectionheading{} command. Note the need to keep any \label that
% is to refer to the section immediately after \section in the above as
% \IEEEraisesectionheading puts \section within a raised box.

% The very first letter is a 2 line initial drop letter followed
% by the rest of the first word in caps (small caps for compsoc).
%
% form to use if the first word consists of a single letter:
% \IEEEPARstart{A}{demo} file is ....
%
% form to use if you need the single drop letter followed by
% normal text (unknown if ever used by the IEEE):
% \IEEEPARstart{A}{}demo file is ....
%
% Some journals put the first two words in caps:
% \IEEEPARstart{T}{his demo} file is ....
%
% Here we have the typical use of a "T" for an initial drop letter
% and "HIS" in caps to complete the first word.

\IEEEPARstart{C}{onvolutional}  neural networks (CNNs) have become the dominant machine learning approach for visual object recognition. Although they were originally introduced over 20 years ago~\cite{lecun}, improvements in computer hardware and network structure have enabled the training of  truly deep CNNs only recently. The original LeNet5~\cite{lenet5} consisted of 5 layers, VGG featured 19~\cite{vgg}, and thanks to the skip/shortcut connections, Highway Networks~\cite{highway} and Residual Networks (ResNets)~\cite{resnet} have surpassed the 100-layer barrier.

As CNNs become increasingly deep, a new research problem emerges: information about the input or gradient that passes through many layers it can vanish and ``wash out'' by the time it reaches the end (or beginning) of the network.
Many recent publications address this problem. For example, Rectified Linear Unites (ReLU)~\cite{relu} avoid gradient saturation, batch-normalization~\cite{batch-norm} reduces covariate shift across layers by re-scaling the outputs of its previous layer.
ResNets~\cite{resnet} and Highway Networks~\cite{highway} bypass signal from one layer to the next via identity connections. Stochastic depth~\cite{stochastic} shortens ResNets by randomly dropping layers during training to allow better information and gradient flow.
FractalNets \cite{fractalnet} repeatedly combine several parallel layer sequences with different number of convolutional blocks to obtain a large nominal depth, while maintaining many short paths in the network.
Although these different approaches vary in network topology and training procedure, they all share a key characteristic: they create short paths from early layers to later layers.

In this paper, we propose an architecture that distills this insight into a simple connectivity pattern: to ensure maximum information flow between layers in the network, we connect \emph{all layers} (with matching feature-map sizes) directly with each other. To preserve the feed-forward nature, each layer obtains additional inputs from all preceding layers and passes on its own feature-maps to all subsequent layers.
\figurename~\ref{fig:ccnn} illustrates this layout schematically.
Crucially, in contrast to ResNets, we never combine features through summation before they are passed into a layer; instead, we combine features through concatenations.
Hence, the $\ell^{th}$ layer has $\ell$ inputs, consisting of the feature-maps of all preceding convolutional blocks. Its own feature-maps are passed on to all $L-\ell$ subsequent layers. This introduces $\frac{L(L+1)}{2}$  connections in an $L$-layer network, instead of just $L$, as in traditional architectures.
Because of its dense connectivity pattern, we refer to our approach as {\emph{\methodnamecap{} (\methodnameshort{})}.

\begin{figure}[t]
%\vspace{-2 ex}
      \centering
      \includegraphics[width=0.5\textwidth]{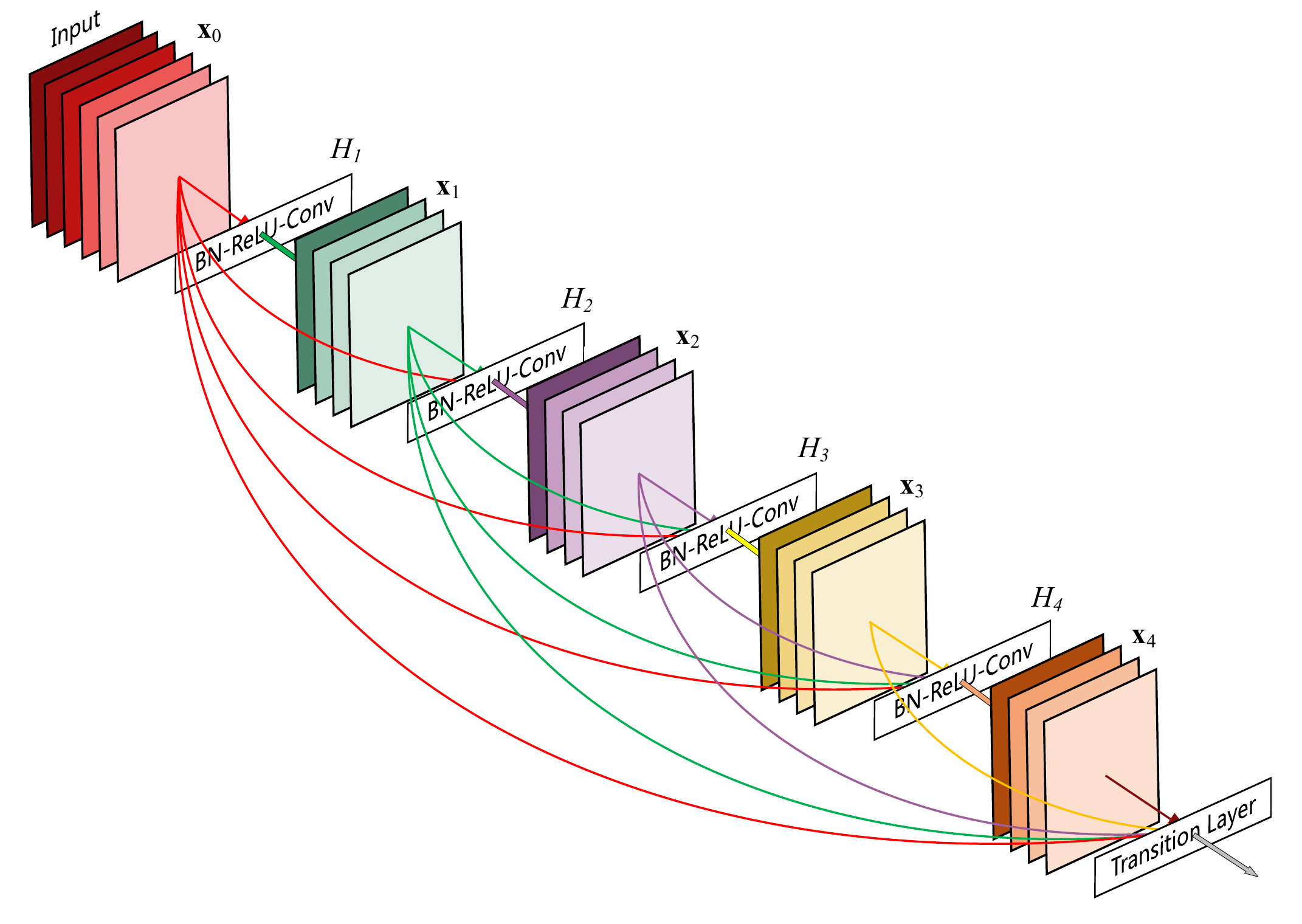}
      \caption{A 5-layer \methodblock{} with a growth rate of $k=4$. Each layer takes all preceding feature-maps as input. }
      \label{fig:ccnn}
      \vspace{-2 ex}
\end{figure}

A possibly counter-intuitive effect of this dense connectivity pattern is that it requires \emph{fewer} parameters than traditional convolutional networks, as there is no need to re-learn redundant feature maps.
Traditional feed-forward architectures can be viewed as algorithms with a state, which is passed on from layer to layer. Each layer reads the state from its preceding layer and writes to the subsequent layer. It changes the state but also passes on information that needs to be preserved. ResNets~\cite{resnet} make this information preservation explicit through additive identity transformations.
Recent variations of ResNets~\cite{stochastic} show that many layers contribute very little and can in fact be randomly dropped during training. This makes the state of ResNets similar to (unrolled) recurrent neural networks~\cite{liao2016bridging}, but the number of parameters of ResNets is substantially larger because each layer has its own weights.
Our proposed \methodnameshort{} architecture explicitly differentiates between information that is \emph{added} to the network and information that is \emph{preserved}.
\methodnameshort{} layers are very narrow (\emph{e.g.}, 12 feature-maps per layer), adding only a small set of feature-maps to the ``collective knowledge'' of the network and keeping the remaining feature-maps unchanged---enables the final classifier to base its decision on all feature-maps in the network.

Besides better parameter efficiency, another big advantage of \methodnameshort{}s is that they are easier to train, due to their improved information flow and gradients throughout the network. Each layer has direct access to the gradients from the loss function and the original input signal, facilitating an implicit form of deep supervision \cite{dsn}. Finally, dense connections create many short paths in the network, which have a strong regularizing effect and reduce overfitting on smaller training sets.

We evaluate \methodnameshort{}s on four highly competitive benchmark datasets (CIFAR-10, CIFAR-100, SVHN, and ImageNet). On these, our models exhibit their superior parameter efficiency, and the benefits thereof, in two ways: 1. they tend to require far fewer parameters when compared against alternative algorithms with comparable accuracy; 2. they outperform the current state-of-the-art results on most of the benchmark tasks as the number of model parameters is increased.

The main results of this paper were published originally in its conference version\footnote{ \url{http://openaccess.thecvf.com/content_cvpr_2017/papers/Huang_Densely_Connected_Convolutional_CVPR_2017_paper.pdf}}. However, this longer article provides a more comprehensive analysis and a deeper understanding of the  novel \methodnameshort{} architecture, \emph{e.g.} hyper-parameters (Section~\ref{sec:hyper-params}) and design choices(Section~\ref{sec-variant}). In addition, we also provide detailed instructions on how to implement the model in a memory efficient way (Section~\ref{subsec:memory-efficient} and \ref{subsec:memory-efficient-results}).

%!TEX root=main.tex

\begin{figure*}[t]
  \vspace{-2 ex}
        \centering
        \includegraphics[width=\textwidth]{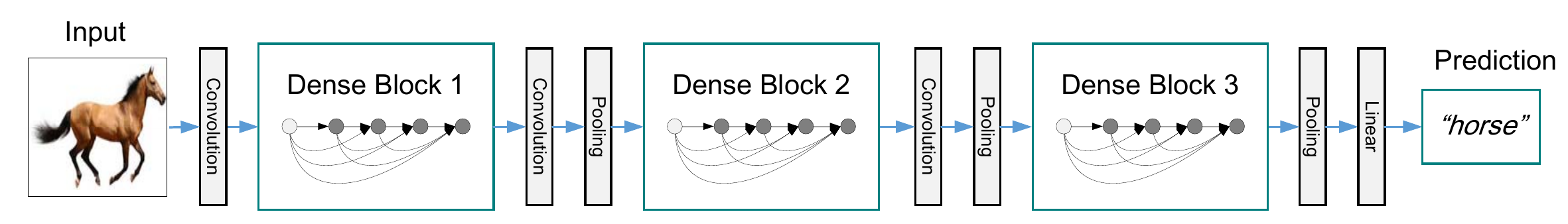}
        \caption{A deep \methodnameshort{} with three dense blocks. The layers between two adjacent blocks are referred to as transition layers and change feature map sizes via convolution and pooling. }
        \label{fig:ccnn_all}
        \vspace{-2 ex}
  \end{figure*}
  
  \section{Related Work}
  The exploration of network architectures has been an integral part of neural network research since their initial discovery.
  The recent resurgence in popularity of neural networks has also revived this research domain.
  The increasing number of layers in modern networks amplifies the differences between architecture types and motivates the exploration of different connectivity patterns as well as the revisiting of old research ideas.
  
  A {cascade structure} similar to our proposed dense connectivity has already been studied in the neural networks literature in the 1980s \cite{fahlman1989cascade}. Their pioneering work focuses on fully connected multi-layer perceptrons trained in a layer-by-layer fashion. More recently, Wilamowski and Yu  \cite{wilamowski2010neural} proposed fully connected cascade networks to be trained with batch gradient descent. Although effective on small datasets, this approach only scales to networks with a few hundred parameters. 
  Several recent publications~\cite{hypercolumn, fcn, pedestrian, yang2015multi}, have found the utilization of multi-level features in CNNs through skip-connections effective for various vision tasks.
  Parallel to our work, \cite{cortes2016adanet} derived a purely theoretical framework for networks with cross-layer connections similar to ours.
  
  Highway Networks \cite{highway}, with their gating units and bypassing paths, were amongst the first architectures that provided a means to effectively train end-to-end networks with more than 100 layers.
  The bypassing paths are presumed to be the key factor that eases the training of these very deep networks, which is further supported by ResNets \cite{resnet}, in which pure identity mappings are used as bypassing paths. ResNets have achieved impressive, record-breaking performance on many challenging image recognition, localization, and detection tasks, such as ImageNet and COCO object detection~\cite{resnet}. Recently, \emph{stochastic depth} was proposed as a way to successfully train a 1202-layer ResNet \cite{stochastic}. Stochastic depth improves the training of deep residual networks by dropping layers randomly during training. This shows that not all layers may be needed and highlights that there is a great amount of redundancy in deep (residual) networks. Our paper was partly inspired by that observation.
  
  An orthogonal approach to making networks deeper (\emph{e.g.}, with the help of skip connections) is to increase the network \emph{width}. The GoogLeNet \cite{szegedy2015going,szegedy2015rethinking} uses an ``Inception module'' which concatenates feature maps produced by filters of different sizes. In \cite{rir}, a variant of ResNets with wide generalized residual blocks was proposed. In fact, simply increasing the number of filters in each layer of ResNets can improve its performance provided the depth is sufficient \cite{wide}. FractalNets also achieve competitive results on several benchmark datasets using a wide network structure \cite{fractalnet}.
  
  Instead of drawing representational power from extremely deep or wide architectures, \methodnameshorts{} exploit the potential of the network through \emph{feature reuse}, yielding condensed models that are easy to train and highly parameter-efficient. Concatenating feature maps learned by \emph{different layers} increases variation in the input of subsequent layers and improves efficiency. This constitutes a major difference between \methodnameshorts{} and ResNets. Compared to Inception networks \cite{szegedy2015going,szegedy2015rethinking}, which also concatenate features from different layers, \methodnameshort{}s are simpler and more efficient.
  
  There are other notable network architecture innovations which have yielded competitive results. The Network in Network (NIN) \cite{netinnet} structure includes micro multi-layer perceptrons into the filters of convolutional layers to extract more complicated features. In Deeply Supervised Network (DSN) \cite{dsn}, internal layers are directly supervised by auxiliary classifiers, which can strengthen the gradients received by earlier layers. Ladder Networks \cite{rasmus2015semi,pezeshki2015deconstructing} introduce lateral connections into autoencoders, producing impressive accuracies on semi-supervised learning tasks. In \cite{wang2016deeply}, Deeply-Fused Nets (DFNs) were proposed to improve information flow by combining intermediate feature representations of different base networks.
  
  Since the original publication of the conference version of this paper, many extensions of \methodnameshort{} have been proposed. These include the Dual Path Network \cite{chen2017dual}, which is a hybrid network architecture combining the \methodnameshort{} and ResNet; the autoencoder \methodnameshort{} \cite{jegou2017one} proposed for semantic segmentation tasks; and 3D-DenseNet used for volume data segmentation \cite{li2017h}. Recently, a number of novel network building modules have been proposed, e.g., group convolution \cite{alexnet}, learned group convolution \cite{huang2017condensenet}, depth-separable convolution \cite{chollet2016xception} and squeeze and excitation \cite{hu2017squeeze}, which are orthogonal innovations to our work and could potentially or have already been shown to be helpful for further improving \methodnameshort{}.

  \section{DenseNets}
  \label{sec:method}
  
  In this section, we first describe the basic design methodology of DenseNet \footnote{Note that unlike its conference version, this paper only presents the DenseNet-BC architecture, i.e., DenseNet with bottleneck layer and transition layer with compression. For notation simplicity, we use the term DenseNet to denote this architecture.}, then we discuss how to implement properly for memory efficient training.
  
  \subsection{The DenseNet Architecture}
  Consider a single image $\bx_0$ that is passed through a convolutional network. The network comprises $L$ layers, each of which implements a non-linear transformation $H_\ell(\cdot)$, where $\ell$ indexes the layer. $H_\ell(\cdot)$ can be a composite function of operations such as Batch Normalization (BN) \cite{batch-norm}, rectified linear units (ReLU)~\cite{relu}, Pooling~\cite{lenet5}, or  Convolution (Conv).
  We denote the output of the $\ell^{th}$ layer as $\bx_{\ell}$.
  % and express the composite transformation in the $\ell^{th}$ layer as $H_\ell(\cdot)$.

  %\vspace{-2 ex}
  \para{ResNets.}
  Traditional convolutional feed-forward networks connect the output of the $\ell^{th}$ layer as input to the $(\ell+1)^{th}$ layer~\cite{alexnet}, which gives rise to the following layer transition: $\bx_\ell=H_\ell(\bx_{\ell-1})$. ResNets~\cite{resnet} add a skip-connection that bypasses the non-linear transformations with an identity function:
  %\vspace{-1ex}
  \begin{equation}
  %\vspace{-1ex}
  \bx_\ell=H_\ell(\bx_{\ell-1})+\bx_{\ell-1}\label{eq:resnet}.
  %%\vspace{-1ex}
  \end{equation}
  An advantage of ResNets is that the gradient flows directly through the identity function from later layers to the earlier layers. However, the identity function and the output of $H_\ell$ are combined by summation, which may cancel out useful features due to the fact the one \emph{cannot} exactly recover the two input features from the summation.
  %impede the information flow in the network.
  
  %\vspace{-2 ex}
  \para{Dense connectivity.}
  To further improve the information flow between layers we propose a different connectivity pattern: we introduce direct connections from any layer to all subsequent layers. \figurename~\ref{fig:ccnn} illustrates the layout of the resulting \methodnameshort{} schematically.
  Consequently, the $\ell^{th}$ layer receives the feature-maps of all preceding layers, $\bx_0,\dots,\bx_{\ell-1}$, as input:
  %\vspace{-1ex}
  \begin{equation}
  %\vspace{-1ex}
  \bx_{\ell} = \ H_\ell([\bx_0, \bx_1,\ldots, \bx_{\ell-1}]),
  \label{eqn:densenet}
  \end{equation}
  where $[\bx_0, \bx_1,\ldots, \bx_{\ell-1}]$ refers to the concatenation of the feature-maps produced in layers $0,\dots,\ell-1$. Because of its dense connectivity we refer to this network architecture as \emph{\methodnamecap{} (\methodnameshort{})}. The recursive concatenation may create massive redundant features in GPU memory during training if not implemented properly. We will discuss this issue in Section~\ref{subsec:memory-efficient}.

  %\vspace{-2 ex}
  \para{Composite function.}
  Following \cite{szegedy2015rethinking,vgg,resnet}, we define $H_\ell(\cdot)$ as a composite function of three types of operations: batch normalization (BN) \cite{batch-norm}, rectified linear unit (ReLU)~\cite{relu} and convolution (Conv). Specifically, each $H_\ell(\cdot)$ corresponds to the sequence: BN-ReLU-Conv(1$\times$1)-BN-ReLU-Conv(3$\times$3). Here, the 1$\times$1 convolution is introduced as a \emph{bottleneck} layer to reduce the number of input feature-maps. This design choice is adopted by many other architectures to improve computational efficiency, and we find it  especially effective for \methodnameshort{}. Unless otherwise specified, each bottleneck layer reduces the input to 4 times the number of feature-maps produced by the subsequent 3$\times$3 convolutional layer. To avoid confusion, we call each $H_\ell$ a \emph{basic layer}, to distinguish it from a single convolutional layer.

  %\vspace{-2 ex}
  \para{Pooling layers.}
  The concatenation operation used in Eq.~(\ref{eqn:densenet}) is not viable when the size of feature-maps changes. However, an essential part of convolutional networks is pooling layers that change the size of feature-maps.  To facilitate pooling in our architecture we divide the network into multiple densely connected \emph{dense blocks}; see \figurename~\ref{fig:ccnn_all}. We refer to layers between blocks as \emph{transition layers}, which do convolution and pooling. The transition layers used in our experiments consist of a batch normalization layer and a 1$\times$1 convolutional layer followed by a 2$\times$2 average pooling layer. Note that the transition layers are much ``wider'' compared to other basic layers, and it is inefficient to use the expensive 3$\times$3 convolution with stride 2 to perform down-sampling in a DenseNet.

  %\vspace{-2 ex}
  \para{Growth rate.}
  If each function $H_\ell$ produces $k$ feature-maps as output, it follows that the $\ell^{th}$ layer has $k\!\times\! (\ell\!-\!1)\!+\!k_0$ input feature-maps, where $k_0$ is the number of channels in the input of that dense block. To prevent the network from growing too wide and to improve the parameter efficiency we limit $k$ to a small integer, \emph{e.g.}, $k=12$. We refer to the hyper-parameter $k$ as the \emph{\stepsizename{}} of the network. We show in Section~\ref{sec:results} that a relatively small \stepsizename{} is sufficient to obtain state-of-the-art results on the datasets that we tested on. One explanation for this is that each layer has access to all the preceding feature-maps in its block and, therefore, to the network's ``collective knowledge''. One can view the feature-maps as the global state of the network. Each layer adds $k$ feature-maps of its own to this state. The  \stepsizename{} regulates how much new information each layer can contribute to the global state.
  The global state, once written, can be accessed from everywhere within the network and, unlike in traditional network architectures, there is no need to replicate it from layer to layer.

  %\vspace{-2 ex}
  \para{Compression.}
  To further improve model compactness, we can reduce the number of feature-maps at transition layers. If a dense block contains $m$ feature-maps, we let the following transition layer generate $\lfloor \theta m\rfloor$ output feature-maps, where $0<\!\theta\le \!1$ is referred to as the compression factor. When $\theta\!=\!1$, the number of feature-maps across transition layers remains unchanged. We set $\theta=0.5$ in our experiments unless otherwise specified.

  \begin{figure*}[t]
    \centering
    \includegraphics[width=0.8\linewidth]{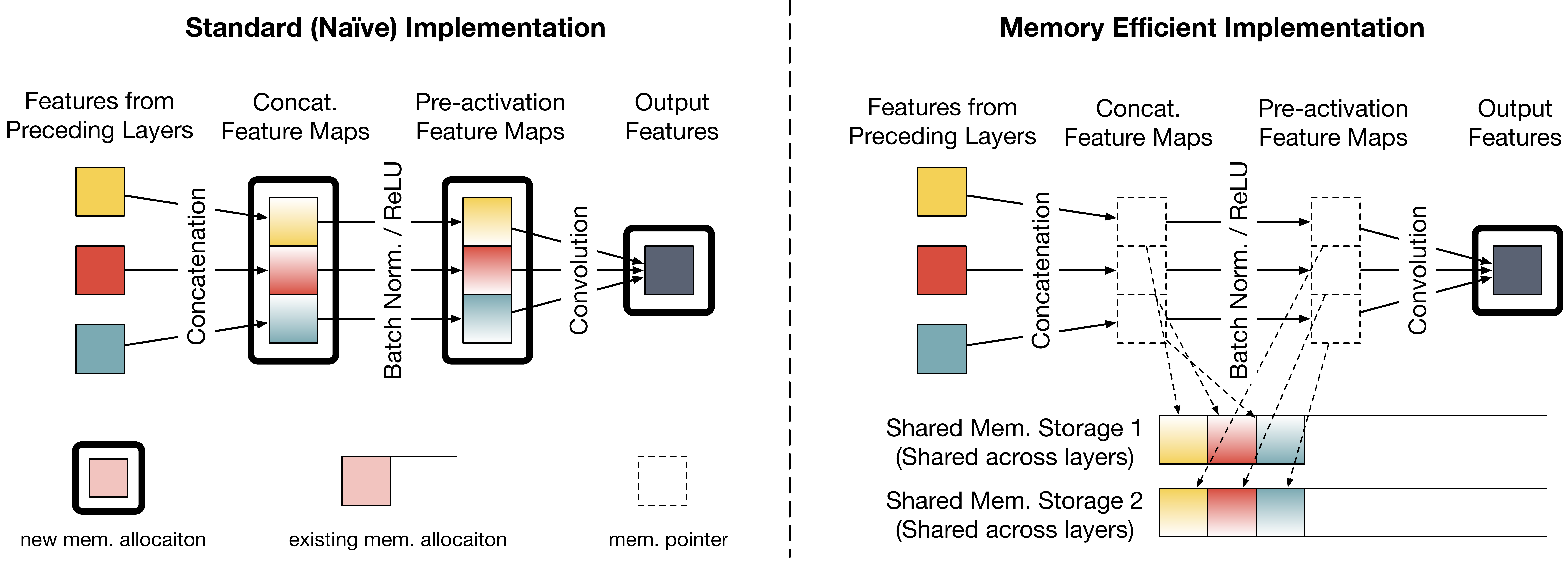}
    \caption{
      DenseNet layer forward pass: original implementation ({\bf left}) and efficient implementation ({\bf right}).
      Solid boxes correspond to tensors allocated in memory, where as translucent boxes are pointers.
      Solid arrows represent computation, and dotted arrows represent memory pointers.
      The efficient implementation stores the output of the concatenation and pre-activation batch normalization/ReLU operations in temporary storage buffers, whereas the original implementation allocates new memory.
    }
    \label{fig:forward}
      \vspace{-3ex}
  \end{figure*}

  \subsection{Implementation Details}
  \label{subsec:memory-efficient}
  DenseNet is a novel architecture that most deep-learning frameworks have not yet been optimized for. If implemented naively, deep-learning frameworks may copy feature maps in every concatenation, producing many redundant copies of the same feature maps. In turn, this may lead to prohibitively high memory consumption on GPU during training. In this subsection, we discuss how to implement DenseNet correctly, and demonstrate that good implementations of DenseNets are in fact very memory-efficient.
  
  The DenseNet computation graph is illustrated in the left-hand side of \figurename~\ref{fig:forward} (left) for a simplified (pre-activation, no bottleneck) layer $H_\ell$ during training.
  This computation graph shows the components of the layer's composite function: 1) previous features are concatenated, 2) a pre-activation batch normalization/ReLU non-linearity are applied, and then 3) a convolution operation produces the next output feature.
  (The ReLU non-linearity is computed in-place, and therefore we combine it with batch normalization for simplicity.)
  We refer to the outputs of the concatenation and pre-activation batch normalization as the \emph{intermediate feature maps} and the new convolutional feature as the \emph{output feature}. Each basic layer $H_\ell$ produces $k$ feature maps, and an $M$-layer dense block has $k\!\times\! M$ output features.

  \subsubsection{Memory-Efficient Implementation of DenseNets}
  \label{subsec:naive_implementation}
  %\para{The standard (na\"ive) DenseNet implementation.}
  During \emph{inference}, an $M$-layer dense block stores $\bigo{M}$ output features in memory for use by the transition layer or final classifier.
  The intermediate feature maps are only used to compute their respective output feature and do not need to be stored.
  The GPU memory usage is therefore \emph{linear} in the depth of the network. Although DenseNet needs to store the output features of multiple layers during inference, it still requires less memory during inference as its layers
  are very ``narrow''. For example, inferencing a ResNet-50 needs to store $256$ feature maps of size $56\times56$ at each layer in its first stage; while the 121-layer DenseNet needs to store at most $64/2+6\times 32 = 224$ feature maps of the same size \footnote{This corresponds to the first dense block. Later dense blocks require less memory as their feature maps are of lower resolution, although the number of channels is larger.}.
  
  %Most implementations will additionally store all of the DenseNet's parameters in GPU memory, however, the number of parameters is less than that of comparably performing networks.
  
  During \emph{training}, the network needs the intermediate feature maps not only for computing output features but also for computing parameter gradients.
  Most deep learning libraries will store all the intermediate feature maps in GPU memory until the forward and backward passes are complete. Obviously, if new space is allocated to store the concatenated features at each layer, the outputs of the $l$th layer have $M\!-\!l\!+\!1$ copies in the memory, leading to a rapid growth in memory consumption.
  
  To avoid this redundancy, we can pre-allocate a single memory buffer that will ultimately contain all the output feature maps of a dense block. The computation of a single layer involves reading the relevant feature maps from the shared memory buffer, computing the outputs of the layer, and storing those outputs in a consecutive part of the memory buffer. In tensor libraries that support operations on strided tensors (such as cudnn), all these operations can be performed in-place, which leads to a memory-efficient implementation of the feature-maps logic without requiring the complex memory management that is required to make other convolutional network architectures memory-efficient. We illustrate the memory pattern in \figurename~\ref{fig:forward}. We implemented it in LuaTorch, and found it allows us to reduce the memory usage by more than $2 \times$ compared to a naive implementation.

  \subsubsection{Further Reducing Memory Usage}
  \label{subsec:mem-efficient}
  Although sharing the memory of concatenated features could avoid saving redundant output features, the pre-activation batch normalization at each layer still needs store a normalized copy of all the previous output features. This also accounts for a quadratic memory consumption with respect to the network depth. Using post-activation batch normalization could circumvent this problem, while it generally leads to significant worse results (see Section~\ref{subsec:post-act}).
  
  Fortunately, the batch normalization layer (and the subsequent ReLU) is much cheaper to compute compared to convolution. Instead of storing all feature maps for the backward pass, one could \emph{recompute} normalized feature maps on-the-fly when they are needed for gradient computations. Therefore, we only need to allocate a global memory that are shared by all the batch normalization layers (later BN layers simply override the outputs of earlier ones).
  This strategy can be generally applied to other architectures (as shown in \cite{chen2016training}), while it is especially useful for DenseNet, as it allows us to train DenseNets with very small memory consumption.
  With this optimization, we are able to train three times larger models using the same amount of memory, while introducing little computation time overhead. Additional results are presented in Section~\ref{subsec:memory-efficient-results}.

  \renewcommand{\arraystretch}{1.1}
  \begin{table*}[!t]
  % Please add the following required packages to your document preamble:
  % \usepackage{multirow}
  \centering
  \caption{DenseNet architectures for ImageNet. The growth rate for all the networks is $k=32$. Note that each ``conv'' layer shown in the table corresponds to the sequence BN-ReLU-Conv, except that the first``conv'' layer with filter size $7\!\times\!7$ corresponds to Conv-BN-ReLU.}
  \resizebox{0.8\textwidth}{!}{%
  \begin{tabular}{c|c|c|l|l|l}
  \hline
  Layers    & Output Size     & DenseNet-121        & \multicolumn{1}{c|}{DenseNet-169}             & \multicolumn{1}{c|}{DenseNet-201}             & \multicolumn{1}{c}{DenseNet-265}               \\ \hline
  Convolution             & \cross{112} & \multicolumn{4}{c}{\cross{7} conv, stride 2}             \\ \hline
  Pooling                 & \cross{56}   & \multicolumn{4}{c}{\cross{3} max pool, stride 2}         \\ \hline
  \begin{tabular}[c]{@{}c@{}}Dense Block\\ (1)\end{tabular}                       & \cross{56}   & \multicolumn{1}{l|}{\conv{6}}  & \conv{6}  & \conv{6}  & \conv{6} \\ \hline
  \multirow{2}{*}{\begin{tabular}[c]{@{}c@{}}Transition Layer\\ (1)\end{tabular}} & \cross{56}  & \multicolumn{4}{c}{\cross{1} conv}                       \\ \cline{2-6}
       & \cross{28}   & \multicolumn{4}{c}{\cross{2} average pool, stride 2}            \\ \hline
  \begin{tabular}[c]{@{}c@{}}Dense Block\\ (2)\end{tabular}                       & \cross{28}   & \multicolumn{1}{l|}{\conv{12}} & \conv{12}& \conv{12} & \conv{12} \\ \hline
  \multirow{2}{*}{\begin{tabular}[c]{@{}c@{}}Transition Layer\\ (2)\end{tabular}} & \cross{28}  & \multicolumn{4}{c}{\cross{1} conv}                       \\ \cline{2-6}
       & \cross{14}   & \multicolumn{4}{c}{\cross{2} average pool, stride 2}\\ \hline
  \begin{tabular}[c]{@{}c@{}}Dense Block\\ (3)\end{tabular}                       & \cross{14}   & \multicolumn{1}{l|}{\conv{24}} & \conv{32} & \conv{48} & \conv{64} \\ \hline
  \multirow{2}{*}{\begin{tabular}[c]{@{}c@{}}Transition Layer\\ (3)\end{tabular}} & \cross{14}   & \multicolumn{4}{c}{\cross{1} conv}                   \\ \cline{2-6}
     & \cross{7}     & \multicolumn{4}{c}{\cross{2} average pool, stride 2}           \\ \hline
  \begin{tabular}[c]{@{}c@{}}Dense Block\\ (4)\end{tabular}                       & \cross{7}   & \multicolumn{1}{l|}{\conv{16}} & \conv{32} & \conv{32} & \conv{48}  \\ \hline
  \multirow{2}{*}{\begin{tabular}[c]{@{}c@{}}Classification\\ Layer\end{tabular}} & \cross{1} & \multicolumn{4}{c}{\cross{7} global average pool}                  \\ \cline{2-6}
    &                 & \multicolumn{4}{c}{1000D fully-connected, softmax}            \\ \hline
  \end{tabular}
  }
  %\vspace{1 ex}
  \label{densenet-imagenet}
  \vspace{-2 ex}
  \end{table*}

  %\vspace{-2 ex}
  \subsection{Network Configurations}
  On the CIFAR and SVHN datasets, the \methodnameshort{}s used in our experiments have 3 dense blocks that each have an equal numbers of layers. Before entering the first dense block, a 3$\times$3 convolution with $2\!\times\! k$ filters is performed on the input images.
  By varying the number of basic layers ($M$) in each block, we can create models with different depth ($L$)\footnote{Following existing works, network depth corresponds to the number of layers with trainable weights, \emph{e.g.}, convolutional layers and fully connected layers. However, batch normalization layers are not counted.}. It is easy to see that $L\!=\!6\!\times\!M\!+\!4$.
  For convolutional layers with kernel size 3$\times$3, each side of the inputs is zero-padded by one pixel to keep the feature-map size unchanged. We use 1$\times$1 convolution followed by 2$\times$2 average pooling as transition layers between two contiguous dense blocks. At the end of the last dense block, a global average pooling is performed and then a softmax classifier is attached. The feature-map sizes in the three dense blocks are 32$\times$ 32, 16$\times$16, and 8$\times$8, respectively. We experimented with \stepsizename{} and layer configurations $\{L\!=100, k\!=\!12\}$, $\{L\!=\!250,k\!=\!24\}$ and $\{L\!=\!190,k\!=\!40\}$.
  
  On ImageNet, we use a structure with 4 dense blocks on 224$\times$224 input images. The initial convolution layer comprises 2$\times k$ convolutions of size 7$\times$7 with stride 2; the number of feature-maps in all other layers is a function of $k$. The exact network configurations we used on ImageNet are shown in Table~\ref{densenet-imagenet}.

%%%%%%%%%%%%%%%%%%%%%%%%%%%%%%%%%%%%%%%%%%%%%%%%%%%%%%%%%%%%%%%%%%%%%%%%%%%%%%%%%%%%%%%%
%%%%%%%%%%%%%%%%%%%%%%%%%%%%%%%%%%%%%%%%%%%%%%%%%%%%%%%%%%%%%%%%%%%%%%%%%%%%%%%%%%%%%%%%
%%%%%%%%%%%%%%%%%%%%%%%%%%%%%%%%%%%%%%%%%%%%%%%%%%%%%%%%%%%%%%%%%%%%%%%%%%%%%%%%%%%%%%%%

\section{Results}
\label{sec:results}

\newcommand{\resultcaption}{Error rates (\%) on CIFAR and SVHN datasets. $L$ denotes the network depth and $k$ its \stepsizename{}. Results that surpass all competing methods are {\textbf{bold}} and the overall best results are {\color{blue}\textbf{blue}}. ``+'' indicates standard data augmentation (translation and/or mirroring). $\ast$ indicates results run by ourselves. All the results of \methodnameshort{}s without data augmentation (C10, C100, SVHN) are obtained using Dropout. \methodnameshort{}s achieve lower error rates while using fewer parameters than ResNet. 
Without data augmentation, \methodnameshort{} performs better by a large margin.
}
% Please add the following required packages to your document preamble:
% \usepackage{graphicx}
\renewcommand{\arraystretch}{1.0}
\begin{table*}[]
\centering
\small
\caption{\resultcaption}
\resizebox{0.7\textwidth}{!}{%
\begin{tabular}{l|cc|cc|cc|c}
\hline
\multicolumn{1}{c|}{Method} & \multicolumn{1}{l}{Depth} & \multicolumn{1}{l|}{Params} & C10 & C10+ & C100 & C100+ & SVHN \\ \hline
Network in Network \cite{netinnet} & - & - & 10.41 & 8.81 & 35.68 & - & 2.35 \\
All-CNN \cite{allcnn} & - & - & 9.08 & 7.25 & - & 33.71 & - \\
Deeply Supervised Net \cite{dsn} & - & - & 9.69 & 7.97 & - & 34.57 & 1.92 \\
Highway Network \cite{highway} & - & - & - & 7.72 & - & 32.39 & - \\ \hline
FractalNet \cite{fractalnet} & 21 & 38.6M & 10.18 & 5.22 & 35.34 & 23.30 & 2.01 \\
with Dropout/Drop-path & 21 & 38.6M & 7.33 & 4.60 & 28.20 & 23.73 & 1.87 \\ \hline
ResNet \cite{resnet} & 110 & 1.7M & - & 6.61 & - & - & - \\ \hline
ResNet (reported by \cite{stochastic}) & 110 & 1.7M & 13.63 & 6.41 & 44.74 & 27.22 & 2.01 \\ \hline
ResNet with Stochastic Depth \cite{stochastic}  & 110 & 1.7M & 11.66 & 5.23 & 37.80 & 24.58 & 1.75 \\
 & 1202 & 10.2M & - & 4.91 & - & - & - \\ \hline
Wide ResNet \cite{wide} & 16 & 11.0M & - & 4.81 & - & 22.07 & - \\
 & 28 & 36.5M & - & 4.17 & - & 20.50 & - \\
with Dropout & 16 & 2.7M & - & - & - & - & 1.64 \\ \hline
ResNet (pre-activation) \cite{identity-mappings} & 164 & 1.7M & 11.26$^\ast$ & 5.46 & 35.58$^\ast$ & 24.33 & - \\
\multicolumn{1}{c|}{} & 1001 & 10.2M & 10.56$^\ast$ & 4.62 & 33.47$^\ast$ & 22.71 & - \\ \hline
\methodnameshort{} $(k=12)$ & 100 & 0.8M & \textbf{5.92} & 4.51 & \textbf{24.15} & 22.27 & 1.76 \\
\methodnameshort{} $(k=24)$ & 250 & 15.3M & {\color{blue}\textbf{5.19}} & \textbf{3.62} & {\color{blue}\textbf{19.64}} & \textbf{17.60} & 1.74 \\
\methodnameshort{} $(k=40)$ & 190 & 25.6M & - & {\color{blue}\textbf{3.46}} & - & {\color{blue}\textbf{17.18}} & - \\ \hline
\end{tabular}%
}
\vspace{0.5ex}
\label{my-label}
\vspace{-2ex}
\end{table*} 
\label{sec:experiment-results}
  We empirically demonstrate the effectiveness of \methodnameshort{} on several benchmark datasets
  and compare it with state-of-the-art network architectures, especially with ResNet and its variants.
  %Code is available at \url{https://github.com/liuzhuang13/DenseNet}.

\subsection{Datasets}
%\vspace{-1 ex}
\para{CIFAR.}
      The two CIFAR datasets \cite{cifar} consist of colored natural scene images, with 32$\times$32 pixels each. CIFAR-10 (C10) consists of images drawn from 10 and CIFAR-100 (C100)  from 100 classes. The train and test sets contain 50,000 and 10,000 images respectively and we hold out 5,000 training images as a validation set.
      We adopt a standard data augmentation scheme that is widely used for this dataset \cite{resnet, stochastic, fractalnet, netinnet, fitnet, dsn, allcnn, highway}: the images are first zero-padded with 4 pixels on each side, then randomly cropped to again produce 32$\times$32 images; half of the images are then horizontally mirrored. We denote this augmentation scheme by a ``+'' mark at the end of the dataset name (\emph{e.g.}, C10+). For data preprocessing, we normalize the data using the channel means and standard deviations. We evaluate our algorithm on all four datasets: C10, C100, C10+, C100+. For the final run we use all 50,000 training images and report the final test error at the end of training.

\para{SVHN.}
    The Street View House Numbers (SVHN) dataset \cite{svhn} contains 32$\times$32 colored digit images coming from Google Street View. The task is to classify the central digit into the correct one of the 10 digit classes. There are 73,257 images in the training set, 26,032 images in the test set, and 531,131 images for additional training. Following common practice~\cite{maxout, stochastic, dsn, netinnet, sermanet2012convolutional} we use all the training data without any data augmentation, and split a validation set with 6,000 images from the training set. We select the model with the lowest validation error during training and report the test error. We follow \cite{wide} and divide the pixel values by 255 so they are in the $[0, 1]$ range.

\para{ImageNet.} The ILSVRC 2012 classification dataset \cite{deng2009imagenet} consists 1.2 million images for training, and 50,000 for validation. Each image is associated with a label from $1,000$ predefined classes. We adopt the same data augmentation scheme for the training images as in \cite{blogresnet,resnet,identity-mappings}, and apply a 224$\times$224 center crop to images at test time. Following common practice \cite{resnet,identity-mappings,stochastic}, we report classification errors on the validation set.

%\vspace{-2 ex}
\subsection{Training}
All the networks are trained using SGD. On CIFAR and SVHN we train using mini-batch size 64 for 300 and 40 epochs, respectively. The initial learning rate is set to 0.1, and is divided by 10 at 50\% and 75\% of the total number of training epochs. On ImageNet, we train models for 90 epochs\footnote{Training for more epochs on ImageNet generally leads to higher accuracy. But for a fair comparison with ResNets, we limit the training of DenseNets to 90 epochs.} with a mini-batch size of 256. The learning rate is set 0.1 initially, and is lowered by a factor of 10 after epoch 30 and epoch 60.

Following \cite{blogresnet}, we use a weight decay of $10^{-4}$ and a Nesterov momentum \cite{nesterov} of 0.9 without dampening. We adopt the weight initialization introduced by \cite{init}. For the three datasets without data augmentation, \emph{i.e.}, C10, C100 and SVHN, we add a dropout layer \cite{dropout} after each convolutional layer (except the first one) and set the dropout rate to 0.2. The test errors were only evaluated once for each task and model setting.

\begin{figure*}[t]
\centerline{
\resizebox{\textwidth}{!}{
    \includegraphics[width=0.28\linewidth]{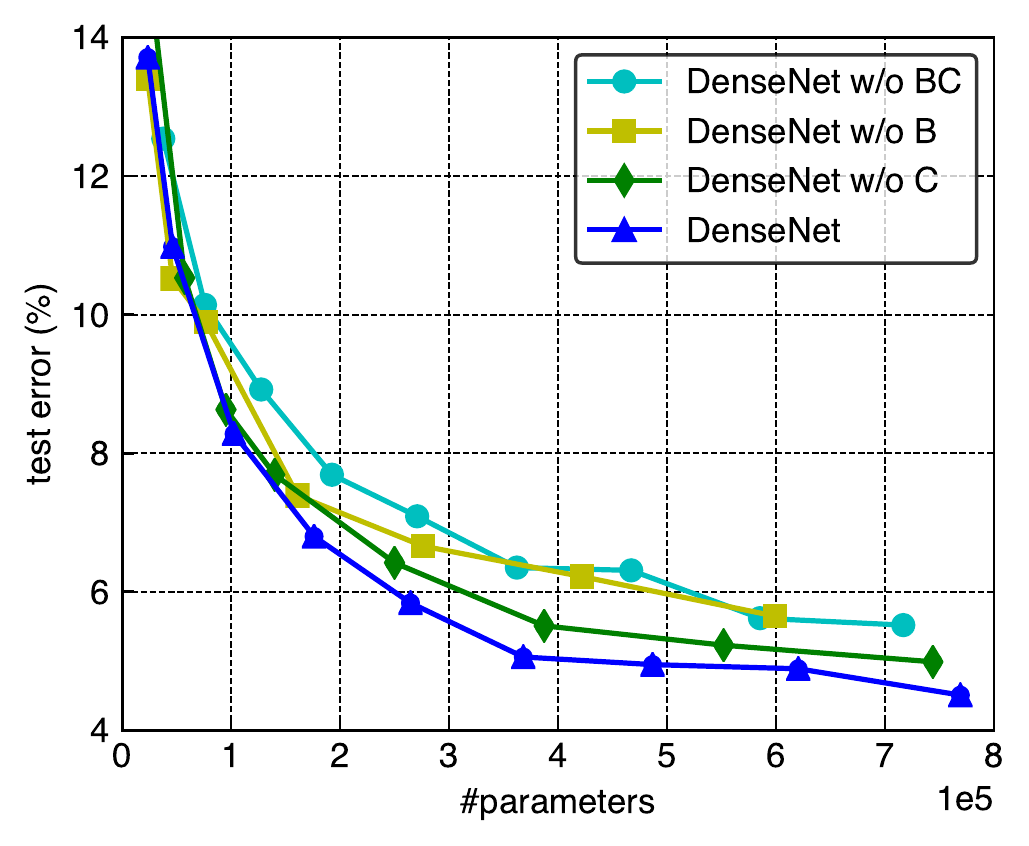}
    \includegraphics[width=0.28\linewidth]{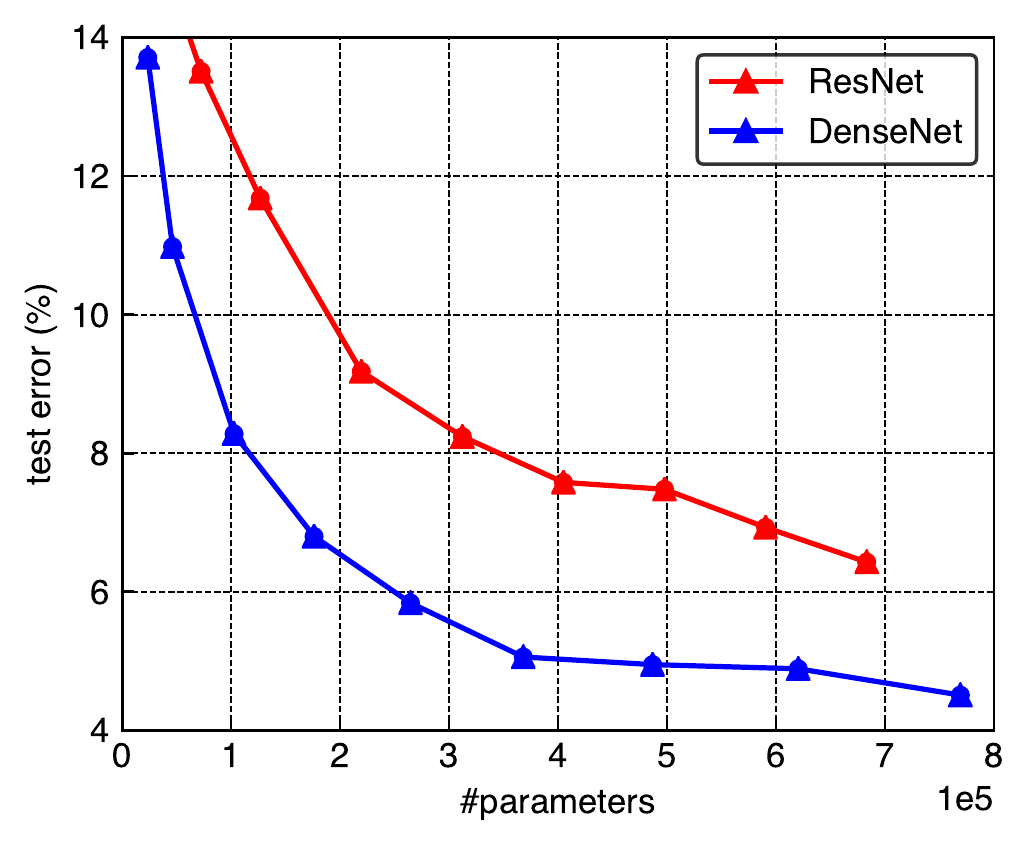}
      \includegraphics[width=0.406\textwidth]{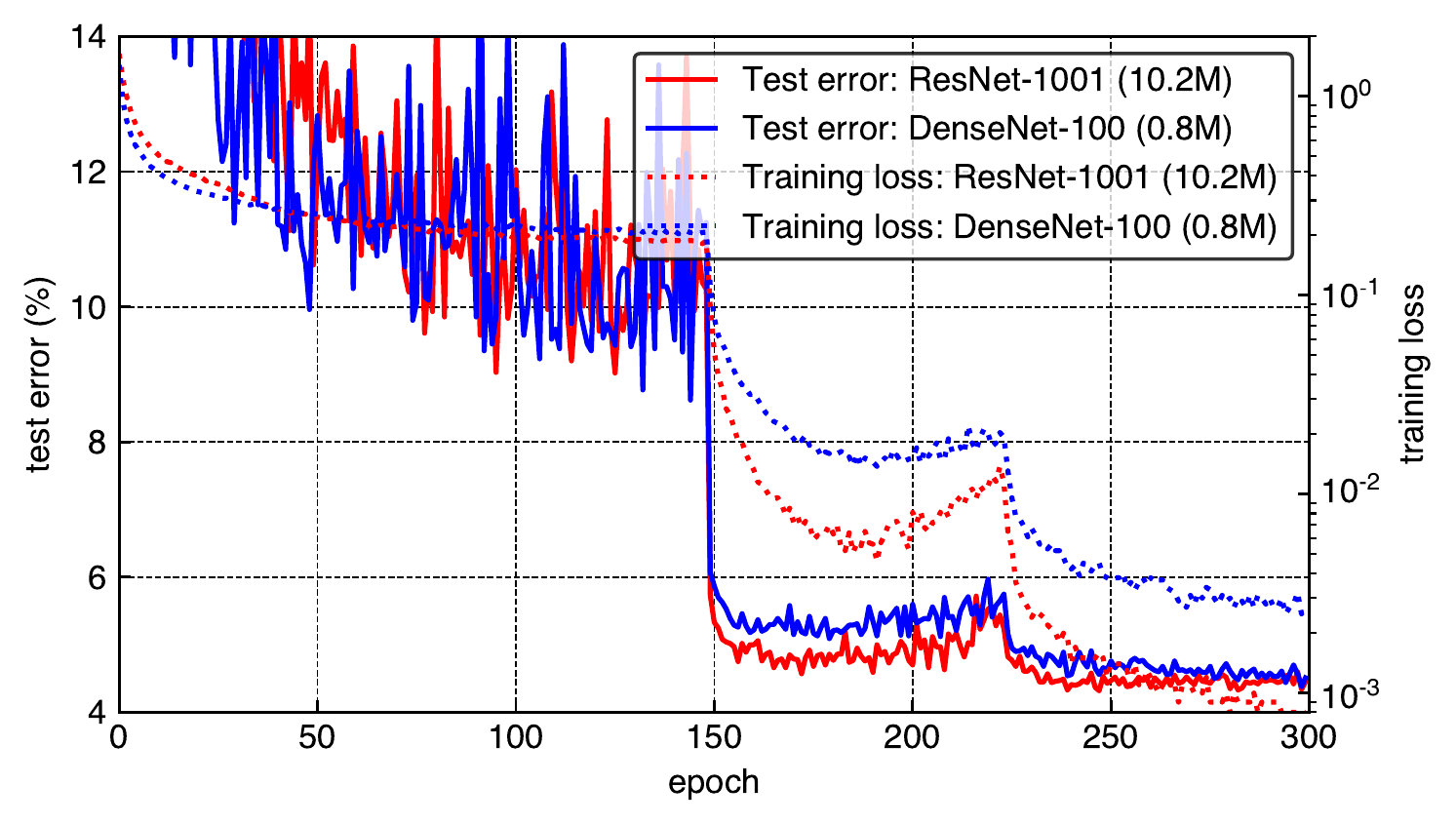}}
    }
      \caption{\emph{Left:} Comparison of the parameter efficiency on C10+ between \methodnameshort{} variations. \emph{Middle:} Comparison of the parameter efficiency between \methodnameshort{} and (pre-activation) ResNets. \methodnameshort{} requires about 1/3 of the parameters as ResNet to achieve comparable accuracy.
      \emph{Right:} Training and testing curves of the 1001-layer pre-activation ResNet \cite{identity-mappings} with more than 10M parameters and a 100-layer \methodnameshort{} with only 0.8M parameters.}
      \label{fig:params}
      \vspace{-1ex}
\end{figure*}

\subsection{Classification Results on CIFAR and SVHN}

We train \methodnameshort{}s with different depths, $L$, and \stepsizename{}s, $k$.
The main results on CIFAR and SVHN are shown in Table~\ref{my-label}.
To highlight general trends, we mark all results that outperform the existing state-of-the-art  in {\textbf{boldface}} and the overall best result in {\color{blue} \textbf{blue}}.

\para{Accuracy.}
Possibly the most noticeable trend is observable in the bottom row of Table~\ref{my-label}, which shows that \methodnameshort{} with $L\!=\!190$ and $k\!=\!40$ outperforms the existing state-of-the-art consistently on \emph{all} the CIFAR datasets. Its error rates of 3.46\% on C10+ and 17.18\% on C100+ are significantly lower than the error rates achieved by wide ResNet architecture~\cite{wide}. Our best results on C10 and C100 (without data augmentation) are even more encouraging: both are close to 30\% lower than FractalNet with drop-path regularization \cite{fractalnet}. On SVHN, with dropout, the \methodnameshort{} with $L\!=\!100$ and $k\!=\!24$ also surpasses the current best result achieved by wide ResNet. However, the 250-layer DenseNet does not further improve the performance significantly over its shorter counterpart. This may be explained by the fact that SVHN is a relatively easy task and very deep models tend to overfit to the training set.

\para{Capacity.} Without compression or bottleneck layers, there is a general trend that \methodnameshorts{}  perform better as $L$ and $k$ increase. We attribute this primarily to the corresponding growth in model capacity. This is best demonstrated by the column of C10+ and C100+. On C10+,  the error drops from 4.51\% to 3.62\% and finally to 3.46\% as the number of parameters increases from 0.8M, over 15.3M to 25.6M. On C100+, we observe a similar trend. This suggests that \methodnameshorts{} can utilize the increased representational power of bigger and deeper models. It also indicates that DenseNets do not suffer from overfitting or the optimization difficulties of residual networks~\cite{resnet}.

%\vspace{-2 ex}
\para{Parameter Efficiency.}
The results in Table~\ref{my-label} indicate that \methodnameshorts{} utilize parameters more effectively than alternative model architectures (in particular, ResNets). The \methodnameshort{} with bottleneck structure and compression at transition layers is particularly parameter-efficient. For example, our deepest model only has 15.3M parameters, but it consistently outperforms other models such as FractalNet and Wide ResNets with more than 30M parameters. We also highlight that \methodnameshort{} with $L\!=\!100$ and $k\!=\!12$ achieves comparable performance  (\emph{e.g.}, 4.51\% vs 4.62\% error on C10+, 22.27\% vs 22.71\% error on C100+) as the 1001-layer pre-activation ResNet using 9$\times$ fewer parameters.
\figurename~\ref{fig:params} (right panel) shows the training loss and test error of these two networks on C10+. The 1001-layer deep ResNet converges to a lower training loss value but a similar test error. We analyze this effect in more detail below.

%\vspace{-2 ex}
\para{Overfitting.}  One positive side-effect of the more efficient use of parameters is a tendency of \methodnameshorts{} to be less prone to overfitting.
%We did observe that typically Dropout is not necessary with \methodnameshort{} architectures, although we did use it for datasets without data augmentation (which are substantially smaller and where overfitting is more of a nuisance).
We observe that on the datasets without data augmentation, the improvements of \methodnameshort{} architectures over prior work are particularly pronounced. On C10, the improvement denotes a 29\% relative reduction in error from 7.33\% to 5.19\%. On C100, the reduction is about 30\% from 28.20\% to 19.64\%.

\begin{figure*}
  \centering
  \includegraphics[width=0.8\textwidth]{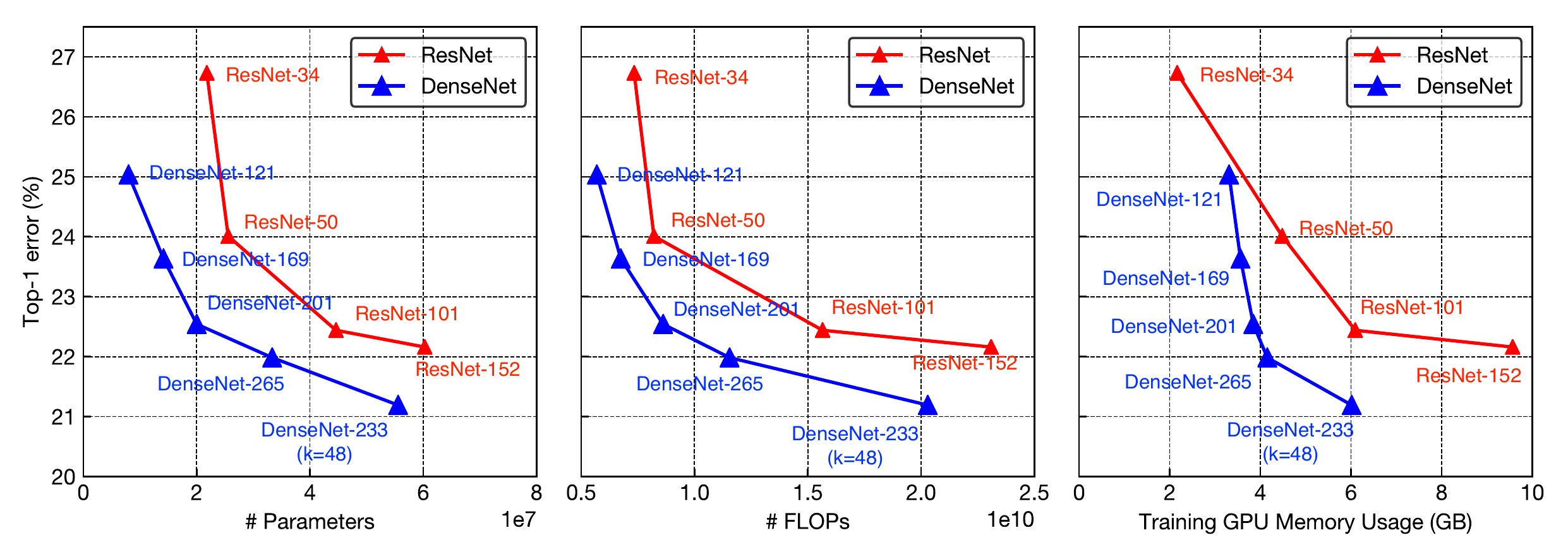}
    % \vspace{-2ex}
  \caption{Comparison of the DenseNet and ResNet Top-1 (single model and single-crop) error rates on the ImageNet classification dataset as a function of learned parameters (\emph{left}), flops (\emph{middle}, and GPU memory footprint at training time(\emph{right}). Training GPU memory measured using the efficient LuaTorch DenseNet implementation with a batch size of $64$.}
  \label{fig:imagenet}
  % \vspace{-3ex}
\end{figure*}

\begin{table}
  \centering
\resizebox{0.7\columnwidth}{!}{
    \begin{tabular}[b]{c|c|c}\hline
      Model & top-1 & top-5 \\ \hline
      \methodnameshort-121& 25.02 / 23.61 & 7.71 / 6.66\\
      \methodnameshort-169& 23.80 / 22.08 & 6.85 / 5.92\\
      \methodnameshort-201& 22.58 / 21.46 & 6.34 / 5.54\\
      \methodnameshort-264& 22.15 / 20.80 & 6.12 / 5.29\\ \hline
    \end{tabular}
    }
    \caption{The top-1 and top-5 error rates on the ImageNet validation set, with single-crop / 10-crop testing.}
    \label{table:imagenet-numbers}
  \vspace{-3ex}
\end{table}

\subsection{Classification Results on ImageNet}
We evaluate \methodnameshort{} with different depths and growth rates on the ImageNet classification task, and compare it with state-of-the-art ResNet architectures. To ensure a fair comparison between the two architectures, we eliminate all other factors such as differences in data preprocessing and optimization settings by adopting the publicly available LuaTorch implementation for ResNet \cite{blogresnet}\footnote{\url{https://github.com/facebook/fb.resnet.torch}}.
We simply replace the ResNet model with the \methodnameshort{} network and keep all the experimental settings unchanged to match those used for ResNets. All results were obtained with single centered test-image crops.
\figurename~\ref{fig:imagenet} shows the validation errors of \methodnameshorts{} and ResNets on ImageNet as a function of the number of parameters (left) and flops (right). The results presented in the figure reveal that \methodnameshorts{} perform on par with the state-of-the-art ResNets, whilst requiring significantly fewer parameters and computation to achieve comparable performance. For example, a DenseNet-201 with 20M parameters model yields similar validation error as a 101-layer ResNet with more than 40M parameters. Similar trends can be observed from the right panel, which plots the validation error as a function of the number of FLOPs: a \methodnameshort{} that requires as much computation as a ResNet-50 performs on par with a ResNet-101, which requires twice as much computation. It is also clear from these results that DenseNet performance continues to improve measurably as more layers are added.

It is worth noting that our experimental setup implies that we use hyperparameter settings that are optimized for ResNets but not for \methodnameshorts{}. It is conceivable that more extensive hyper-parameter searches may further improve the performance of \methodnameshort{} on ImageNet.

\subsection{Discussion}

Superficially, \methodnameshort{}s are quite similar to ResNets: Eq. (\ref{eqn:densenet}) differs from Eq.~(\ref{eq:resnet}) only in that the inputs to $H_\ell(\cdot)$ are concatenated instead of summed. However, the implications of this seemingly small modification lead to substantially different behaviors of the two network architectures.

\para{Model compactness.}
As a direct consequence of the input concatenation, the feature maps learned by any of the \methodnameshort{} layers can be accessed by all subsequent layers. This encourages feature reuse throughout the network, and leads to more compact models.

The left two plots in \figurename~\ref{fig:params} show the result of an experiment that aims to compare the parameter efficiency of the various variants of \methodnameshorts{} (left) and also a comparable ResNet architecture (middle).
We train multiple small networks with varying depths on C10+ and plot their test accuracies as a function of network parameters.
In comparison with other popular network architectures, such as AlexNet \cite{alexnet} or VGG-net \cite{vgg}, ResNets with pre-activation use fewer parameters while typically achieving better results~\cite{identity-mappings}. Hence, we compare \methodnameshort{} ($k=12$) against this architecture. The training setting for \methodnameshort{} is kept the same as in the previous section.

The graph shows that \methodnameshort{} with bottleneck layer structure and transition layer compression is consistently the most parameter efficient among these variants.
Further, to achieve the same level of accuracy, \methodnameshort{} only requires around 1/3 of the parameters of ResNets (middle plot). This result is in line with the results on ImageNet we presented in \figurename~\ref{fig:imagenet}. The right plot in \figurename~\ref{fig:params} shows that a DenseNet with only 0.8M trainable parameters is able to achieve comparable accuracy as the 1001-layer (pre-activation) ResNet \cite{identity-mappings} with 10.2M parameters.

\para{Implicit Deep Supervision.}
One explanation for the improved accuracy of \methodname{}s may be that individual layers receive additional supervision from the loss function through the shorter connections. One can interpret \methodnameshorts{} to perform a form of ``deep supervision''. The benefits of deep supervision have previously been shown in deeply-supervised nets (DSN; \cite{dsn}), which have classifiers attached to every hidden layer, enforcing the intermediate layers to learn discriminative features.

\methodnameshorts{} perform a similar deep supervision in an implicit fashion: a single classifier on top of the network provides direct supervision to all layers through at most two or three transition layers. However, the loss function and gradient of \methodnameshorts{} are substantially less complicated, as the same loss function is shared between all layers.

\para{Stochastic vs. deterministic connection.} There is an interesting connection between \methodname{}s and stochastic depth regularization of residual networks~\cite{stochastic}. In stochastic depth, layers in residual networks are randomly dropped, which creates direct connections between the surrounding layers. As the pooling layers are never dropped, the network results in a similar connectivity pattern as \methodnameshort{}: There is a small probability for any two layers, between the same pooling layers, to be directly connected---if all intermediate layers are randomly dropped.  Although the methods are ultimately quite different, the \methodnameshort{} interpretation of stochastic depth may provide insights into the success of this regularizer.

\begin{figure}[t]
      \centering
      \includegraphics[width=0.48 \textwidth]{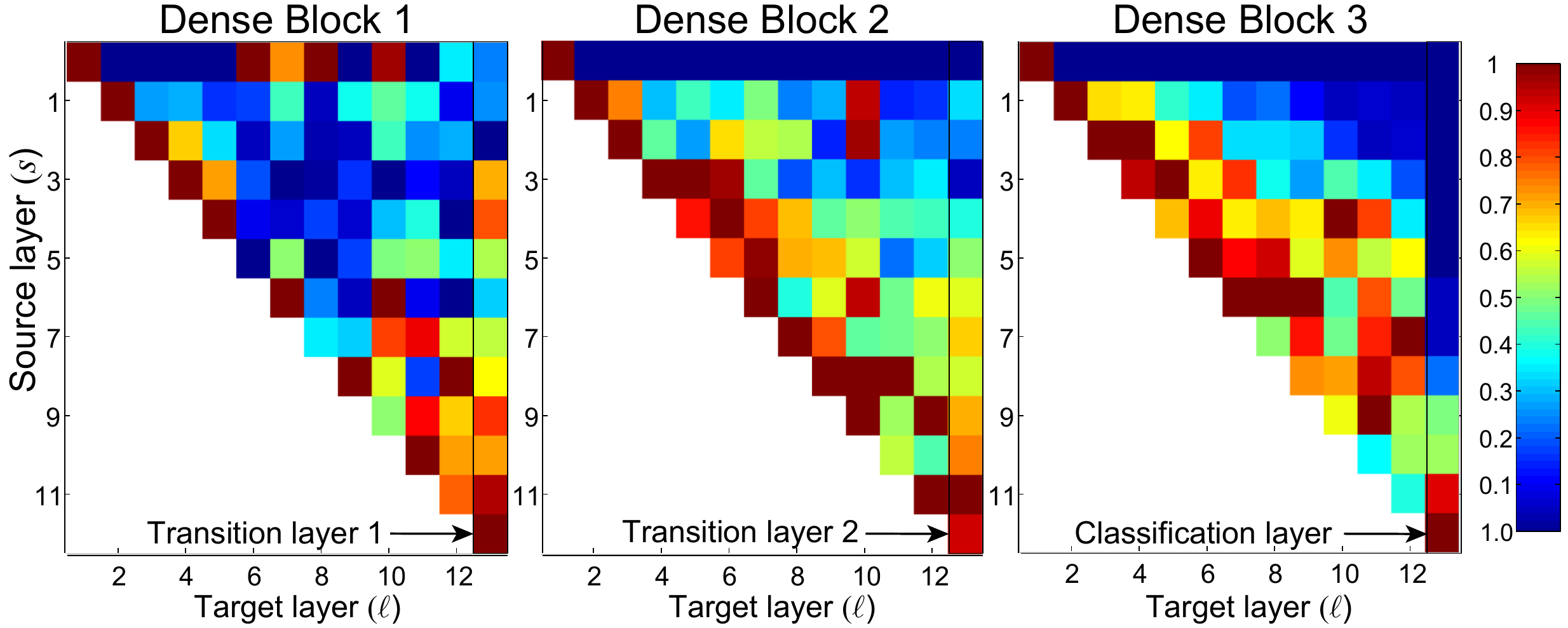}
              % \vspace{-1ex}
      \caption{The average absolute filter weights of convolutional layers in a trained \methodnameshort{}. The color of pixel $(s,\ell)$ encodes the average $L1$ norm (normalized by the number of input feature maps) of the weights connecting convolutional layer $s$ to layer $\ell$ within a dense block. The three columns highlighted by black rectangles correspond to the two transition layers and the classification layer. The first row encodes those weights connected to the input layer of the dense block. }
      \label{fig:weight}
        % \vspace{-3ex}
\end{figure}

\begin{figure*}
  \centering
  \includegraphics[width=0.64\linewidth]{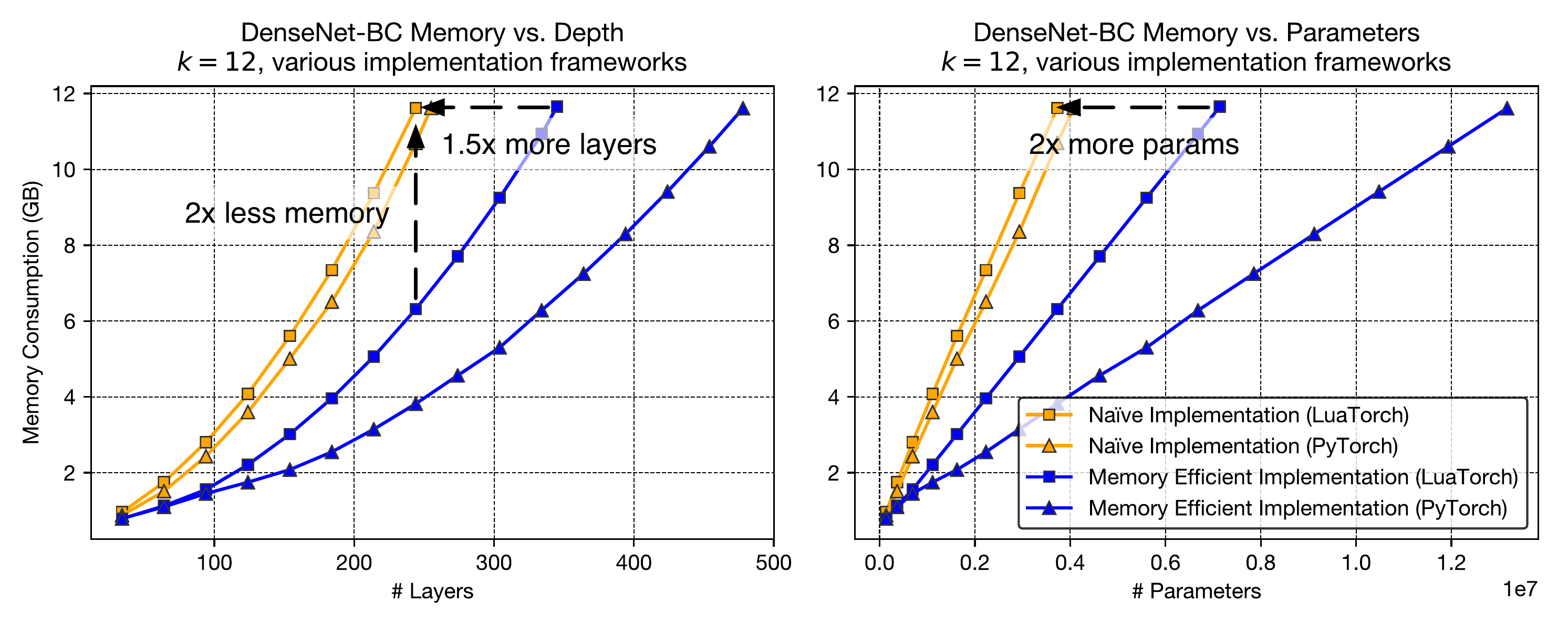}
  \includegraphics[width=.34\linewidth]{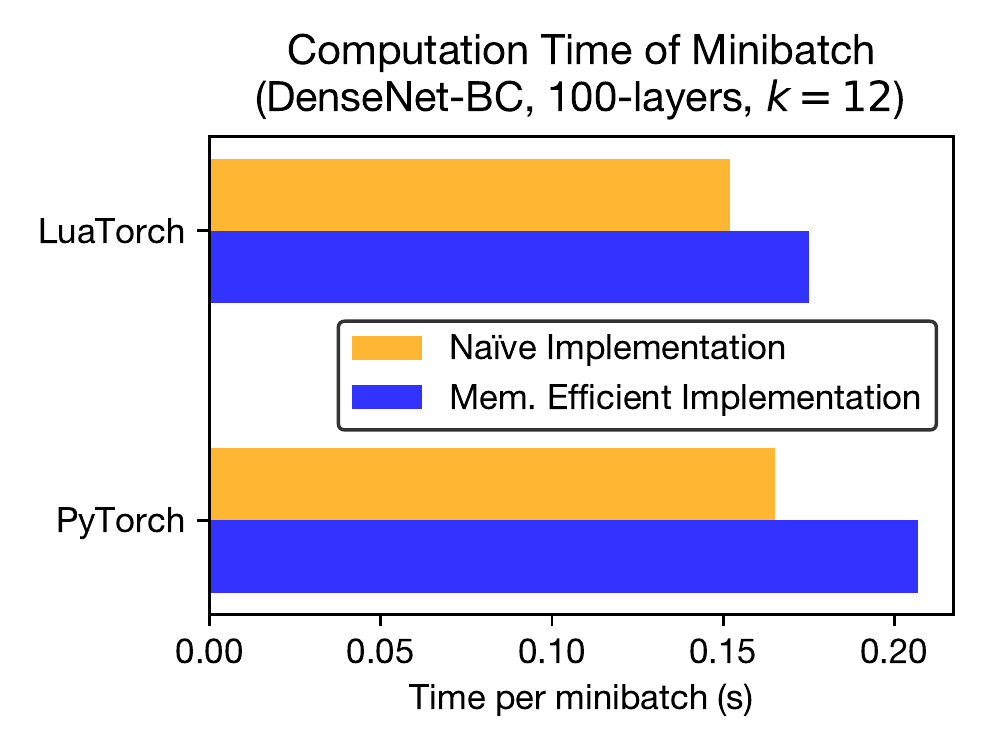}
  \caption{\emph{Left and Middle:}
    GPU memory consumption as a function of network depth/number of parameters.
    Each model is a DenseNet with $k=12$ features added per layer.
    The efficient implementation can train much deeper models with less memory. \emph{Right:} Computation time (on a NVIDIA Maxwell Titan-X).
  }
  \label{fig:memvslayer}
    \vspace{-3ex}
\end{figure*}

\para{Feature Reuse.}
By design, \methodnameshorts{} allow layers access to feature maps from all of its preceding layers (although sometimes through transition layers). We conduct an experiment to investigate if  a trained network takes advantage of this opportunity.
We first train a \methodnameshort{} on C10+ with $L\!=\!40$ and $k\!=\!12$. For each convolutional layer $\ell$ within a block, we compute the average (absolute) weight assigned to connections with layer $s$. \figurename~\ref{fig:weight} shows a heat-map for all three dense blocks.
The average absolute weight serves as a surrogate for the dependency of a convolutional layer on its preceding layers.
A red dot in position ($\ell,s$) indicates that the layer $\ell$  makes, on average, strong use of feature maps produced $s$-layers before.
Several observations can be made from the plot:

\begin{enumerate}
\item All layers spread their weights over many inputs within the same block. This indicates that features extracted by very early layers are, indeed, directly used by deep layers throughout the same dense block.
%\vspace{-1ex}
\item The weights of the transition layers also spread their weight across all layers within the preceding dense block, indicating information flow from the first to the last layers of the \methodnameshort{} through few indirections.
\item The layers within the second and third dense block consistently assign the least weight to the outputs of the transition layer (the top row of the triangles), indicating that the transition layer outputs many redundant features (with low weight on average).  This is in keeping with the strong results of \methodnameshort{} where exactly these outputs are compressed.
\item Although the final classification layer, shown on the very right, also uses weights across the entire dense block, there seems to be a concentration towards final feature-maps, suggesting that there may be some more high-level features produced late in the network.
\end{enumerate}

\begin{figure*}[t]
\centerline{
\resizebox{\textwidth}{!}{
    \includegraphics[width=0.3\linewidth]{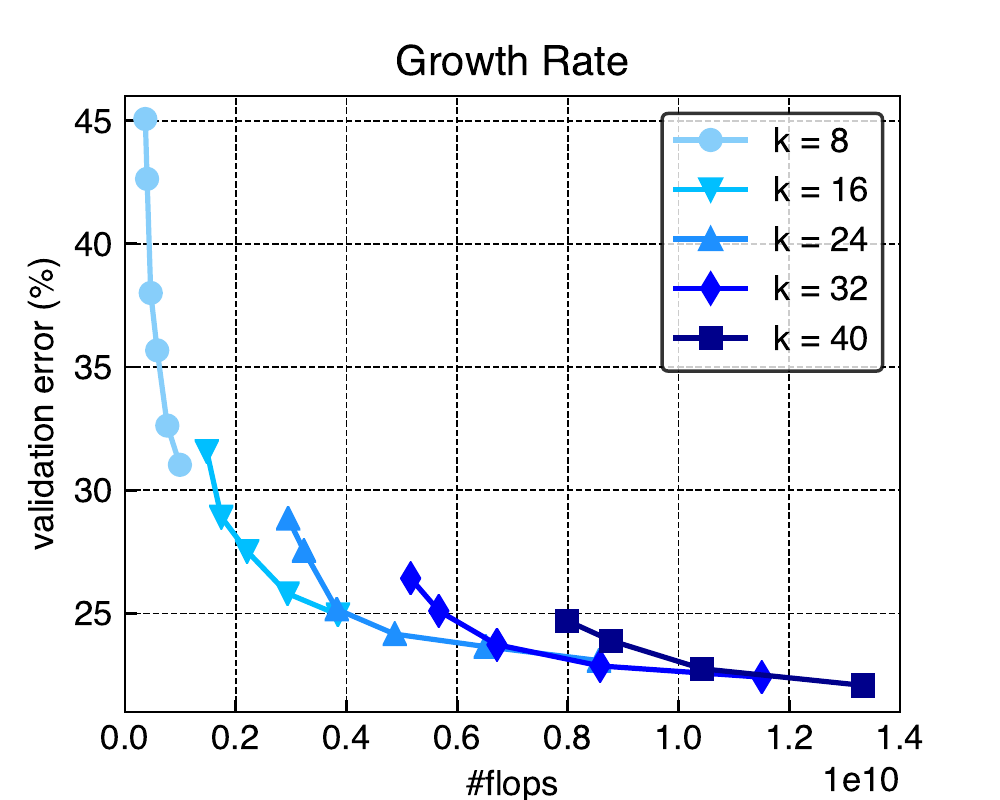}
    \includegraphics[width=0.3\linewidth]{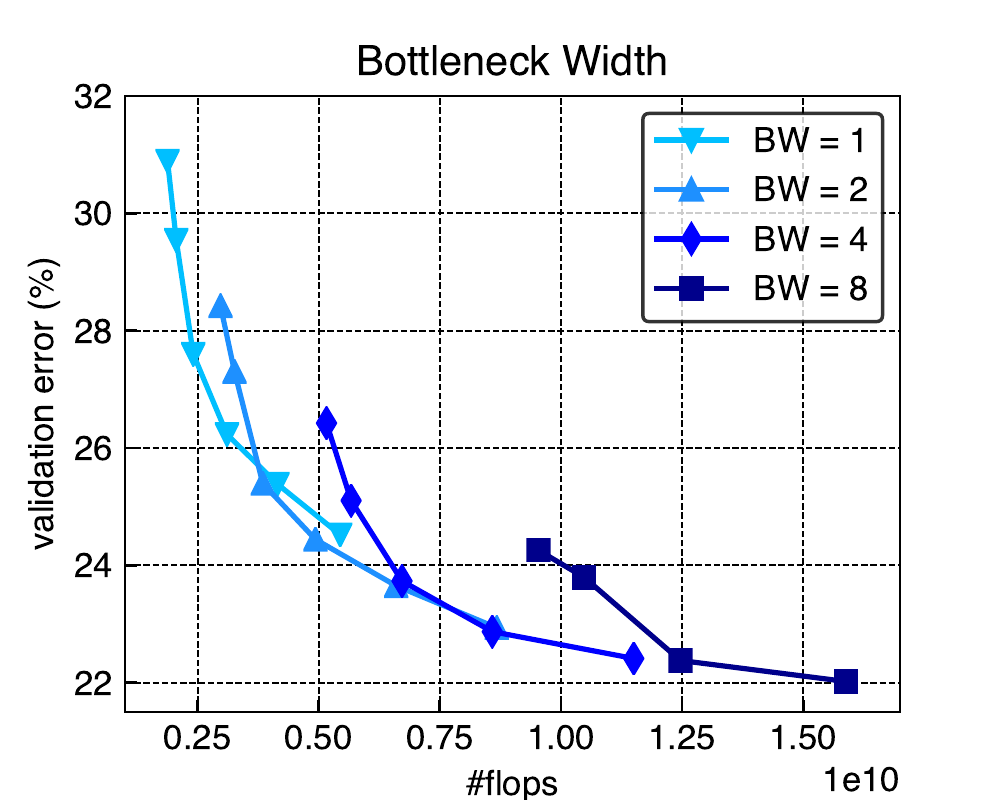}
      \includegraphics[width=0.3\textwidth]{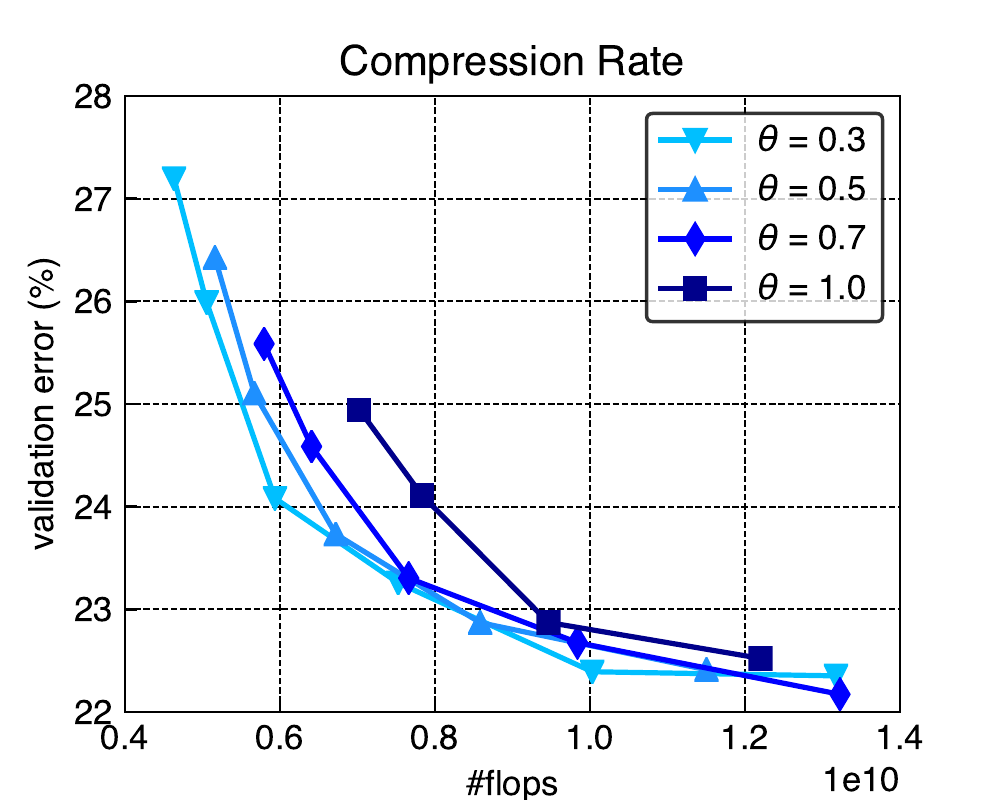}}
    }
          % \vspace{-1ex}
      \caption{Top-1 validation error on ImageNet as a function of the computational cost (measured by flops) of different DenseNets, with varying growth rate (\emph{Left}), varying width of the bottleneck layers (\emph{Middle}) and varying compression ratio at transition layers (\emph{Right}).}
      \label{fig:sensitivity}
      % \vspace{-3ex}
\end{figure*}

\subsection{Memory Efficient Implementation Results}
\label{subsec:memory-efficient-results}

We compare the memory consumption and computation time of three DenseNet implementations during training.
The na\"ive implementation allocates memory for every pre-activation batch normalization operation.
We compare this against a memory-efficient implementation of DenseNets,
which includes all optimizations described in \autoref{subsec:memory-efficient}.
The concatenated and normalized feature maps are recomputed as necessary during back-propagation.
The only tensors stored in memory during training are the convolution feature maps and the parameters of the network.
We test these two implementations in both the LuaTorch and PyTorch\footnote{
  \url{http://github.com/torch/torch7/}, \url{http://github.com/pytorch/pytorch/}
} deep learning frameworks.

\para{Memory consumption.}
We train networks of various depth on the CIFAR-10 dataset.
All networks have a growth rate of $k=12$ and are trained with a batch size of $64$.
In \figurename~\ref{fig:memvslayer}, we see that the na\"ive implementation becomes memory intensive very quickly in both LuaTorch and PyTorch.
The memory usage of a $160$ layer network ($1.8M$ parameters) is roughly $10$ times as much as a $40$ layer network ($160K$ parameters).
Training a larger network with more than $220$ layers requires over 12 GB of memory, pushing the memory limits of a typical single GPU.
On the other hand, using all the memory-sharing operations significantly reduces memory consumption.
In LuaTorch, the $220$-layer model uses $50\%$ of the memory required by the Na\"ive Implementation.
Under the same memory budget (12 GB), it is possible to train a $340$-layer model, which is $1.5\times$  as deep and has $2\times$ as many parameters as the best na\"ive implementation model.
With the PyTorch efficient implementation, we can train $500$ layer networks ($13M$ parameters) on a single GPU.
The ``autograd'' library in PyTorch performs memory optimizations during training, which likely contribute to this implementation's efficiency.

It is worth noting that the \emph{total} memory consumption of the most efficient implementation does not grow linearly with depth, as the number of parameters is inherently a quadratic function of the network depth. This is a function of the architectural design and part of the reason why DenseNets are so efficient. The memory required to store the parameters is far less than the memory consumed by the feature maps and the remaining quadratic term  does not impede model depth.

\para{Training time.}
In the right panel of \figurename~\ref{fig:memvslayer}, we plot the time per minibatch of a 100-layer DenseNet ($k\!=\!12$) on a NVIDIA Maxwell Titan-X.
The efficient implementation adds roughly $15\%$ time overhead on LuaTorch, and $20\%$ on PyTorch.
This extra cost is a result of recomputing the intermediate feature maps during back-propagation.
If GPU memory is limited, the time overhead from sharing batch normalization/concatenation storage constitutes a reasonable trade-off.

\para{ImageNet results.}
We test the new memory efficient LuaTorch implementation on the ImageNet classification dataset.
The deepest model trained using the na\"ive implementation was $201$ layers ($20M$ parameters).
With the efficient LuaTorch implementation however,
we are able to train two deeper DenseNet models with the efficient implementation, one with $265$ layers ($k=32$, $33M$ parameters) and one with $233$ layers ($k=48$, $55M$ parameters).\footnote{
In the $233$-layer model, the four dense blocks have 6, 12, 48, and 48 layers, respectively.
}

\section{Architecture Hyperparameters}
\label{sec:hyper-params}
We perform a series of analytical experiments to study how hyperparameter choices associated with aspects of the network architecture, affect the performance of DenseNets. Specifically, we examine three hyperparameters: the growth rate $k$, the bottleneck width (number of filters in the 1$\times$1 bottleneck layer) and the compression rate at transition layers $\theta$.

As larger models tend to yield higher accuracy, we incorporate the computational cost whenever we compare two models directly. Therefore, we always train multiple DenseNets with varying depth for each hyperparameter setting, and compare different configurations on the error \emph{v.s.} compute (number of \emph{flops}) plot. The depth of the models ranges from 101 layers to 329 layers, with the number of basic layers for the four dense blocks selected from the set $\{[6,12,18,12]$, $[6,12,24,16]$, $[6,12,32,32]$, $[6,12,48,32]$, $[6,12,64,48]$, $ [6,12,80,64]\}$.

\para{Growth rate $k$.}
The growth rate determines the \emph{width} of each layer, i.e., the number of feature maps produced by $H_\ell$ as given in Eq.(\ref{eqn:densenet}). We experiment with various growth rates from the set $\{8, 16, 24, 32, 40\}$.

The results are shown in the left panel of \figurename~\ref{fig:sensitivity}. We can observe that due to the dense connectivity, even DenseNets with very \emph{narrow} layers (e.g., $k\!=\!8$) can be trained effectively. Although each layer only produces 8 feature maps (light dotted blue curve), the model still results in highly competitive results.
In fact, a small growth rate is essential for DenseNets to achieve high computational efficiency. For example, to achieve a 24\% validation error, a DenseNet with growth rate 24 requires about 0.50$\times$10$^{10}$ flops; while a similar architecture with growth rate 40 requires 0.88$\times$10$^{10}$ flops.
However, as the networks grow deeper, larger growth rates seem to show greater potential. This indicates that to achieve high efficiency for a DenseNet, we should ensure its depth and width are compatible.
It is noteworthy that wider convolutional layers can be more efficiently computed on GPUs due to better parallelism. Therefore, a larger growth rate may be preferred if one is more concerned about wall time efficiency during training. %We will discuss this issue in more detail in Section~\ref{subsec-igr}.

\para{Bottleneck layer width.}
The convolution layer with filter size 1$\times$1 introduced to the transformation $H_\ell$ significantly improves the parameter efficiency of DenseNets. It performs dimension reduction on the concatenated feature maps before passing them to the more expensive 3$\times$3 convolution layer. In all our previous experiments, we fix the width of these bottleneck layers to $m\!\times\!k$, where $m\!=\!4$, and $k$ is the growth rate. To understand how the performance of DenseNets is affected by $m$, we train multiple DenseNets with $m$ selected from the set $\{1,2,4,8\}$. All the other hyperparameters are set to their default value, except that we vary the depth of the networks to make the error \emph{v.s.} flops plot.

We show the results in the middle panel of \figurename~\ref{fig:sensitivity}. Here, the trend is similar to what we have observed in the experiment with varying growth rates: wider bottleneck layers (e.g., $m\!=\!8$) yield lower computational efficiency on smaller networks. Therefore, the bottleneck layer width should also be compatible with the network depth in order to maximize the parameter efficiency of DenseNets.

\para{Compression factor.}
In previous experiments with DenseNets, each of the transition layers between two \emph{dense blocks} halves the number of channels, i.e. we have set $\theta=0.5$ throughout.
Here we conduct an experiment to study how sensitive the model performance is to the
compression factor $\theta$.

The results are shown in the right panel of \figurename~\ref{fig:sensitivity}. In general, the parameter efficiency of DenseNets is quite insensitive to the compression rate. There seems to be no significant difference between the three curves with  $\theta=0.3$, $\theta=0.5$ and $\theta=0.7$, when the flops is greater than 0.8$\!\times\!$10$^{10}$. However, we do observe that DenseNets with smaller compression factor consistently outperform those with larger compression factor when model size is small. This is to some extent counterintuitive, as we would expect a small compression factor (more reduction) to be more beneficial in larger models, which tend to create more redundancy in the feature maps due to the quadratically growing connections. A possible explanation is that low level features in deep networks are indeed actively reused by deeper layers, and keeping more information about the input is helpful when the network has more capacity to process.

%    \end{subfigure}
%    \begin{subfigure}[b]{0.24\textwidth}
%        \includegraphics[width=\textwidth]{figures/preact_flops.pdf}
%        \vspace{-4ex}
%        \caption{}
%        \label{subfig:preact_flops}
%    \end{subfigure}
%
%
%    \caption{Top 1 validation error on ImageNet of DenseNet variants. (a)-(b): full dense connectivity; (c)-(d): partial dense connection with fixed maximum span; (e)-(f): partial dense connection with odd-even connectivity patter; (g)-(h): partial dense connection with a logarithmic number of connections; (i)-(j): exponentially increasing growth rate; (k)-(l): post-activation batch normalization. }\label{fig:densenet-variants}
%\end{figure*}

\begin{figure}[t]
  \includegraphics[width=0.24\textwidth]{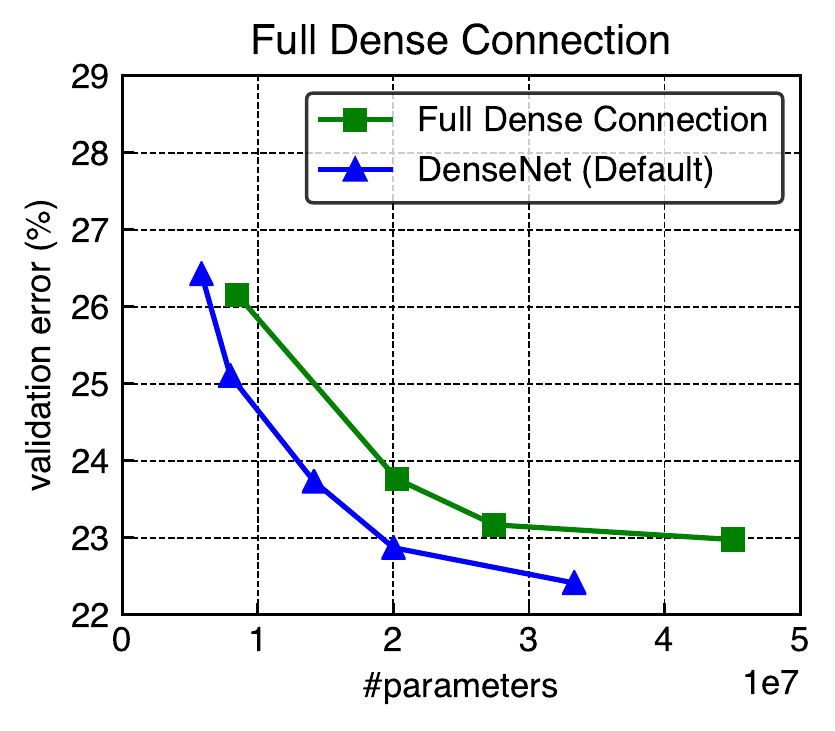}
  \includegraphics[width=0.24\textwidth]{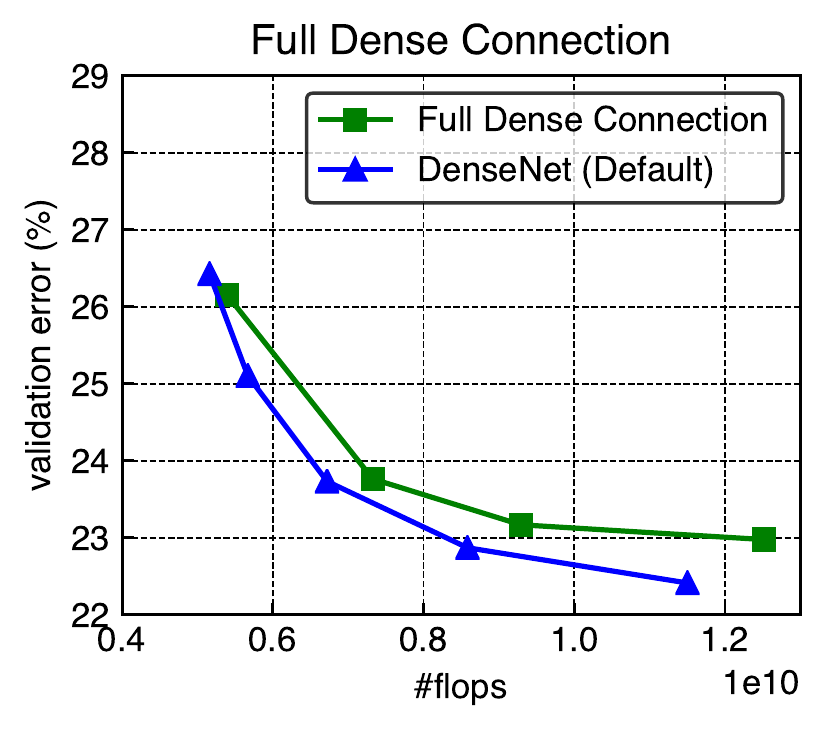}
          \vspace{-2ex}
  \caption{Comparison of DenseNet and its variant with full dense connectivity, in terms of parameter efficiency (\emph{Left}) and computational efficiency (\emph{Right}).}
          \label{fig:fdc}
                  \vspace{-3ex}
\end{figure}

\section{DenseNet variants}
\label{sec-variant}

Multiple variants of the DenseNet architecture are possible, which we discuss and experiment with briefly.
%We empirically investigate several variants of DenseNet to gain a better understanding of the architecture.

\subsection{Full Dense Connectivity}
In a DenseNet, as described so far, layers in different dense blocks are only indirectly connected via a transition layer. This transition layer is necessary because layers in different dense blocks  have incompatible feature map sizes.
One possible variant of the DenseNet layout is to obtain \emph{full dense connectivity} (FDC), and truly connect each layer with every other layer. We can achieve such a configuration by incorporating a  downsampling step via pooling directly into connections between layers with different feature map sizes. This is also equivalent to simplifying the transition layer to a simple pooling operation without $1\!\times\! 1$  convolution and compression.

%layers are directly connected, due to the fact that the pooling operation changes the feature map and makes feature concatenation not viable.
%One interesting question is if is there a way to have layers across different dense blocks densely connected as well? If yes, how does it compare to the previous architecture design?

%A simple idea to have dense connections across layers with different feature sizes is simplifying the transition layer to a single pooling layer. In that way, a deeper layer in the network can directly reuse the down-sampled version of features from earlier stages. We call such design pattern \emph{full dense connectivity} (FDC). Note that FDC is not equivalent to setting the compression rate of transition layers to 1, because the latter has a convolutional operation in the transition layers, which mixes up the feature maps.

We give a comparison of the network with FDC and the original DenseNet in \figurename~\ref{fig:fdc}. It can be observed that DenseNet with transition layers performs significant better in terms of both parameter and computational efficiency. These results suggest that the 1$\!\times$\!1 convolution in transition layers may be helpful in compressing redundant features.

\begin{figure}[t]
  \includegraphics[width=0.24\textwidth]{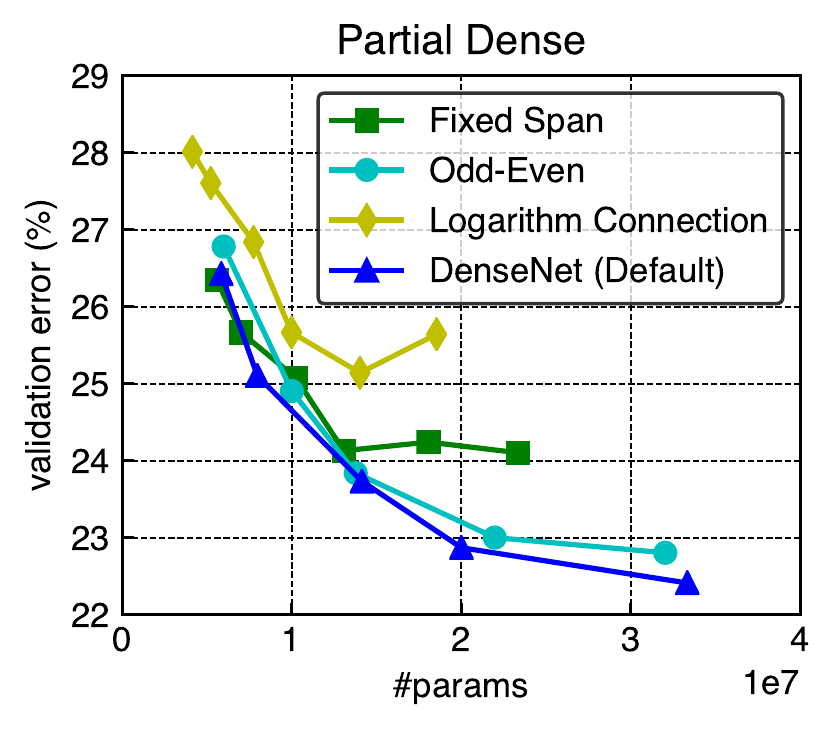}
  \includegraphics[width=0.24\textwidth]{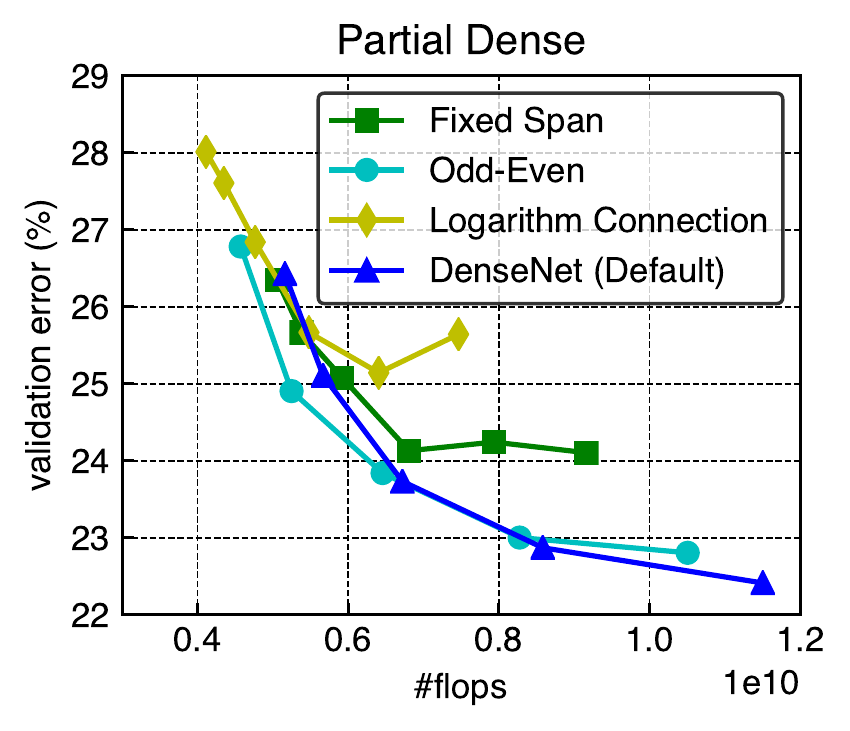}
          \vspace{-2ex}
  \caption{Comparison of DenseNet and its variants with partial dense connectivity, in terms of parameter efficiency (\emph{Left}) and computational efficiency (\emph{Right}).}
  \label{fig:partial}
          \vspace{-3ex}
\end{figure}

\subsection{Partial Dense Connection}
The DenseNet architecture introduces a quadratic number of connections to the network.  It is a fair question to ask if all of these are indeed necessary.
In this subsection, we experiment with some variants of DenseNet, which only keep part of the dense connections. Specifically, we consider three settings: (1) each layer is connected to the most recent $M$ layers; (2) an odd (even) layer is connected to all previous even (odd) layers in that dense block; and (3) each layer is connected to the $2^i$th layer before it\footnote{A similar design  choice was also investigated in \cite{hu2017log}.} , $i=0,1,2,\dots$. Similar to the setting in last subsection, we train multiple networks with varying depth for each architecture and use parameter and computational efficiency as the main  performance criteria.

\figurename~\ref{fig:partial} shows the results of these three architectural variants, as well as the error rate of the standard DenseNet structure. The first variant has dense connections with a span of less than 4, 8 or 12 layers. As shown by the green curve in \ref{fig:partial}, this connectivity pattern tends to yield higher validation error than standard DenseNets. When the depth becomes large, the network cannot be trained effectively and the validation error stops decreasing as the network becomes deeper (the training error follows a similar trend).
These results demonstrate that earlier features are indeed utilized and necessary for deeper layers in a DenseNet, and the long range connections are critical for effectively training very deep models.
The second variant features only half of the number of connections of standard DenseNets. From the cyan curve in  \ref{fig:partial}, one can observe that this connectivity pattern leads to slightly worse parameter efficiency, while it has a higher computational efficiency when depth is not very large. For deeper models this advantage disappears. The third variant networks also under-perform standard DenseNets, especially with larger model sizes. This set of experiments further support our hypothesis that dense connections strengthen the information/gradient flow in the network, and enable training of deep models effectively.

\begin{figure}
\centering
\includegraphics[width=0.24\textwidth]{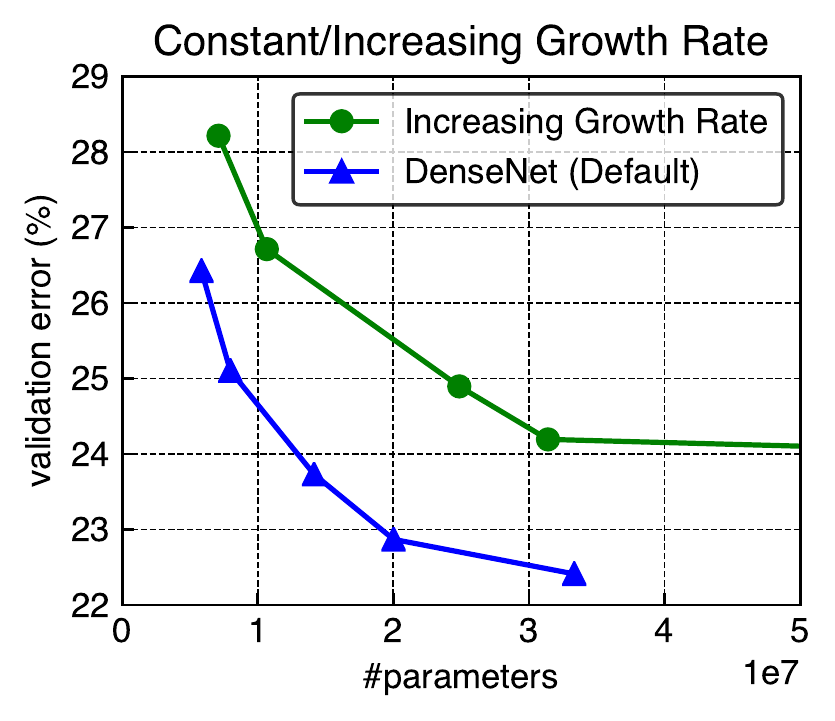}
\includegraphics[width=0.24\textwidth]{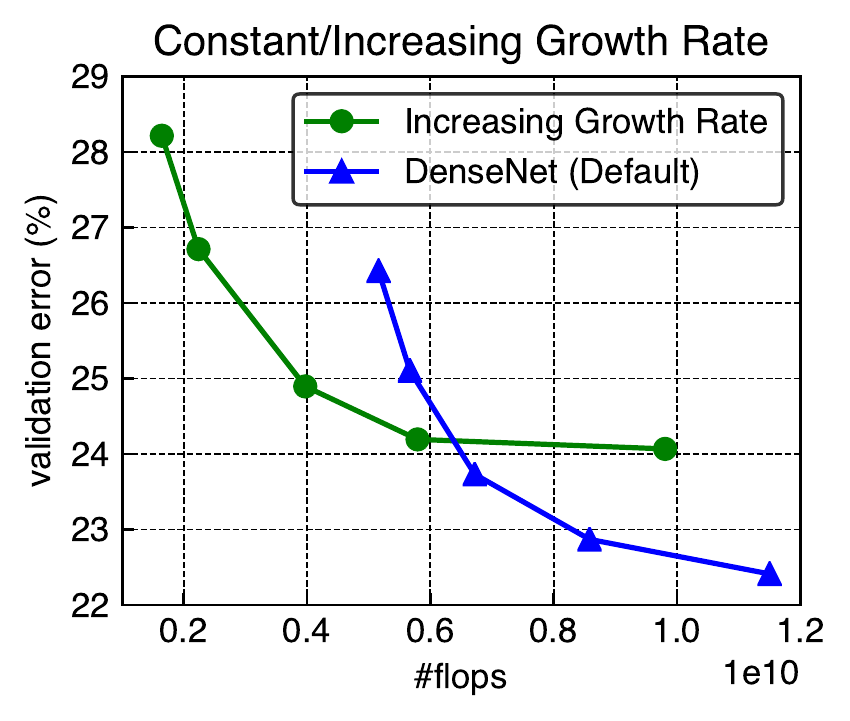}
    \vspace{-2ex}
\caption{Comparison of DenseNet and its variant with increasing growth rate, in terms of parameter efficiency (\emph{Left}) and computational efficiency (\emph{Right}).}\label{fig:doublek}
    \vspace{-3ex}
\end{figure}

\subsection{Exponentially Increasing Growth Rate}
\label{subsec-igr}
%In many real applications, the computational efficiency, usually measured by the number of FLOPs, is of greater interest than the parameter efficiency when we have limited computational resources. Here we perform experiments to explicitly examine the FLOPs efficiency of standard DenseNets and an improved version.

The standard DenseNet use a constant growth rate throughout the whole network. Compared to other network architectures, e.g., the VGG~\cite{vgg} and ResNet~\cite{resnet}, which doubles the number of feature channels after each down-sampling layer, DenseNets/DenseNets tend to allocate fewer parameters towards processing low resolution feature maps. Such a design is oriented more towards parameter efficiency than computational efficiency, as pointed out by~\cite{iandola2016squeezenet}.

In \figurename~\ref{fig:doublek} we compare the standard DenseNet structure with a variant that doubles the growth rate after each transition layer. Standard DenseNets have a constant growth rate of 32, while for the latter network, the growth rate in the $j$th dense block has a growth rate of $k_0\times 2^{j-1}$ ($j=1,2,3,4$).
According to the results, the DenseNet with exponentially increasing growth rate has lower efficiency in terms of parameter efficiency (left panel of \figurename~\ref{fig:doublek}), while it is much more competitive on the error $v.s.$ flops plot (right panel of \figurename~\ref{fig:doublek}). This suggests that using larger growth rate for deeper dense blocks should be preferred in scenarios where computation is the major concern.

\begin{figure}
\centering
\includegraphics[width=0.24\textwidth]{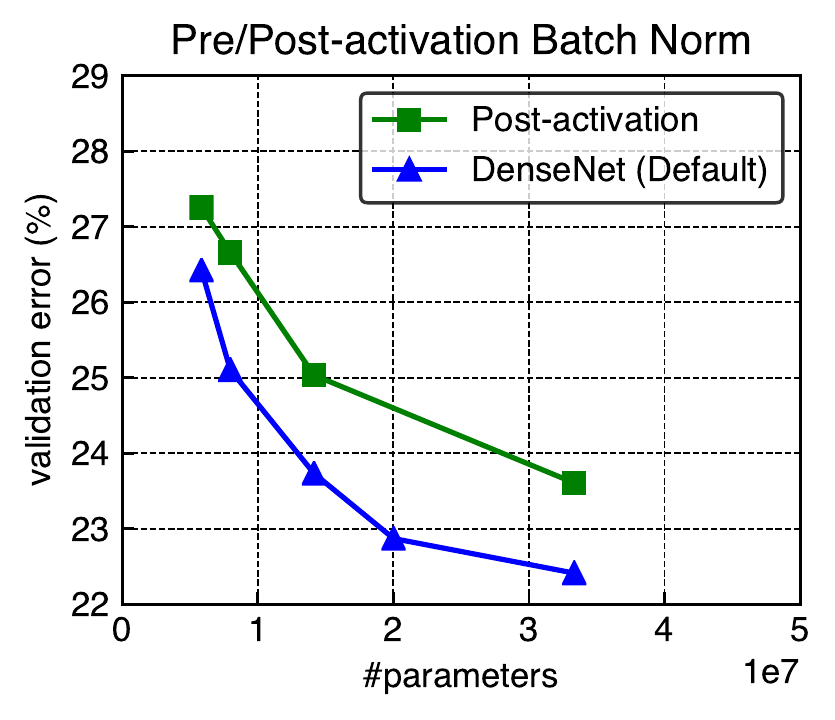}
\includegraphics[width=0.24\textwidth]{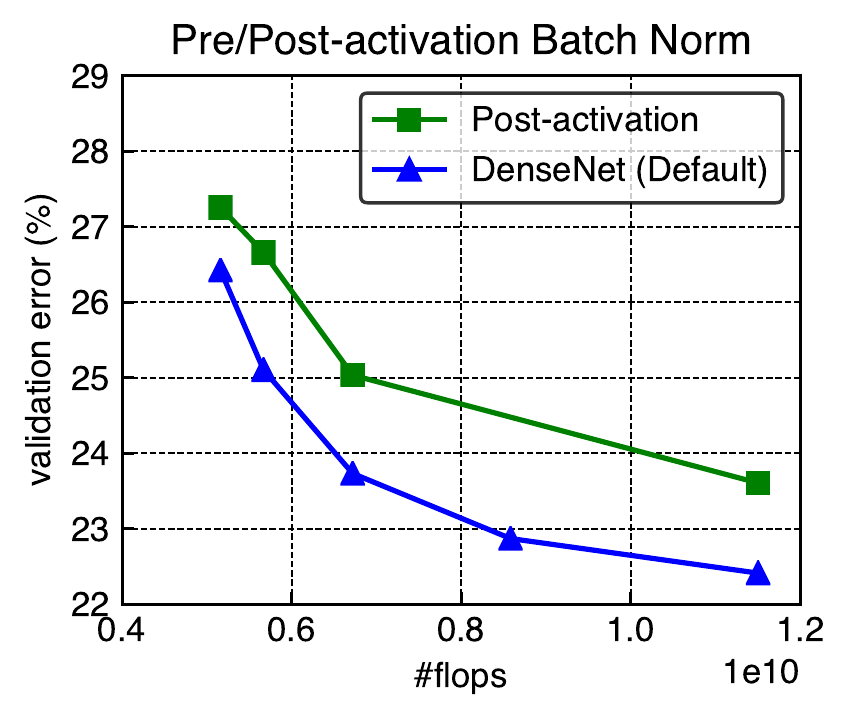}
% \vspace{-2ex}
\caption{Comparison of pre- and post- activation BN in DenseNets, in terms of parameter efficiency (\emph{Left}) and computational efficiency (\emph{Right}).}
\label{fig:preact}
    % \vspace{-3ex}
\end{figure}

\subsection{Post-activation Batch Normalization}
\label{subsec:post-act}
The DenseNet architecture utilizes \emph{pre-activation} batch normalization.
Unlike conventional architectures, pre-activation networks apply batch normalization and non-linearities \emph{before} the convolution operation rather than after.
Though this might seem like a minor change, it makes a big difference in DenseNet performance.
Batch normalization applies a scaling and a bias to the input features.
If each layer applies its own batch normalization operation, then each layer applies a \emph{unique} scale and bias to previous features.
For example, the Layer 2 batch normalization might scale a Layer 1 feature by a constant large than 1, while Layer 3 might scale the same feature by a small positive constant.
This provides Layer 2 and Layer 3 with the flexibility to up-weight and down-weight a same feature independently.
Note that this would not be possible if all layers shared the same batch normalization operation, or if normalization occurred after convolution operations.
\figurename~\ref{fig:preact} provides a comparison between DenseNets with pre- and post- activation batch normalization. There is a clear trend that the former leads to significantly higher efficiency. However, pre-activation batch normalization generally leads to substantially higher memory footprint during training. If GPU memory is limited, we can apply the memory optimization strategy described in Section~\ref{subsec:memory-efficient}.

\section{Conclusion}
We proposed a new convolutional network architecture, which we refer to as \methodnamecap{} (\methodnameshort{}). It introduces direct connections between any two layers with the same feature-map size. Whilst following a simple connectivity rule, \methodnameshorts{} naturally integrate the properties of identity mappings, deep supervision, and diversified depth. They allow feature reuse throughout the network and can consequently learn more compact and, according to our experiments, more accurate models. 

We showed that \methodnameshorts{} scale naturally to hundreds of layers, while exhibiting no optimization difficulties. In our experiments, \methodnameshorts{} tend to yield consistent improvement in accuracy with growing number of parameters, without any signs of performance degradation or overfitting. Under multiple settings, \methodnameshorts{} achieve state-of-the-art results across several highly competitive datasets. Moreover, \methodnameshorts{} require substantially fewer parameters and less computation than prior work at comparable accuracy levels.

\ifCLASSOPTIONcompsoc
  % The Computer Society usually uses the plural form
  \section*{Acknowledgments}
\else
  % regular IEEE prefers the singular form
  \section*{Acknowledgment}
\fi

The authors are supported in part by the NSF III-1618134, III-1526012, IIS-1149882, IIS-1724282, the Office of Naval Research Grant N00014-17-1-2175, the Bill and Melinda Gates foundation, SAP America Inc., and the NSF TRIPODS Award \#1740822 (Cornell TRIPODS Center for Data Science for Improved Decision Making).
We thank Danlu Chen, Daniel Sedra, Tongcheng Li and Yu Sun for many insightful discussions.

% Can use something like this to put references on a page
% by themselves when using endfloat and the captionsoff option.
\ifCLASSOPTIONcaptionsoff
  \newpage
\fi

% trigger a \newpage just before the given reference
% number - used to balance the columns on the last page
% adjust value as needed - may need to be readjusted if
% the document is modified later
%\IEEEtriggeratref{8}
% The "triggered" command can be changed if desired:
%\IEEEtriggercmd{\enlargethispage{-5in}}

% references section

% can use a bibliography generated by BibTeX as a .bbl file
% BibTeX documentation can be easily obtained at:
% http://mirror.ctan.org/biblio/bibtex/contrib/doc/
% The IEEEtran BibTeX style support page is at:
% http://www.michaelshell.org/tex/ieeetran/bibtex/
\bibliographystyle{IEEEtran}
% argument is your BibTeX string definitions and bibliography database(s)
%\bibliography{IEEEabrv,../bib/paper}
\bibliography{citations}

\end{document}

%% file: macros.tex
% METHOD NAME
\newcommand{\methodname}{dense convolutional  network}
\newcommand{\methodnamecap}{Dense Convolutional Network}
\newcommand{\methodnameshort}{DenseNet}
\newcommand{\methodnameshorts}{DenseNets}
\newcommand{\methodblock}{dense block}
\newcommand{\methodblockcap}{Dense Block}

\newcommand{\regmethodname}{feature drop}
\newcommand{\regmethodnamecap}{Feature Drop}

\newcommand{\stepsizename}{growth rate}

\newcommand{\conv}[1]{$\left[\begin{array}{ll} \text{1}\times \text{1} \text{ conv}\\ \text{3}\times \text{3} \text{ conv} \end{array}\right] \times \text{#1}$}

\newcommand{\cross}[1]{#1 $\times$ #1}

% Gradients Equations
\newcommand{\feati}{x_i}
\newcommand{\clsfeati}{y_i}
\newcommand{\featk}{x_k}
\newcommand{\clsfeatk}{y_k}
\newcommand{\loss}{L}
\newcommand{\featL}{x_L}
\newcommand{\clsfeat}{y}
% Inputs and outputs
\newcommand{\anyxs}{\ensuremath{\mathbf{x}}}
\newcommand{\anyys}{\ensuremath{\mathbf{y}}}

\newcommand{\bx}{\ensuremath{\mathbf{x}}}

\newcommand{\sourcexs}{\ensuremath{\mathbf{x^\mathcal{S}}}}
\newcommand{\sourceys}{\ensuremath{\mathbf{y^\mathcal{S}}}}

\newcommand{\targetxs}{\ensuremath{\mathbf{x^\mathcal{T}}}}
\newcommand{\targetys}{\ensuremath{\mathbf{y^\mathcal{T}}}}
\newcommand{\pseudotargetys}{\ensuremath{\mathbf{\hat{y}^\mathcal{T}}}}

\newcommand{\bigo}[1]{\ensuremath{\mathcal O ( #1 )}}